\documentclass{article}
\PassOptionsToPackage{round, compress}{natbib}

\usepackage[preprint]{neurips_2024}

\usepackage[utf8]{inputenc} 
\usepackage[T1]{fontenc}    
\usepackage{hyperref}       
\usepackage{url}            
\usepackage{booktabs}       
\usepackage{amsfonts}       
\usepackage{nicefrac}       
\usepackage{microtype}      
\usepackage{xcolor}         

\usepackage{graphicx}
\usepackage{mathptmx}
\usepackage{multirow}
\usepackage{bookmark}
\usepackage{amsmath}

\usepackage{algorithm}
\usepackage{algpseudocode}
\usepackage{float}
\usepackage{amsthm}

\usepackage{cases}
\usepackage{wrapfig}
\usepackage{subcaption}
\usepackage{bbm}
\usepackage{bm}
\usepackage{threeparttable}
\usepackage{colortbl}
\usepackage{subcaption}

\title{DistPred: A Distribution-Free Probabilistic Inference Method for Regression and Forecasting}

\author{%
Daojun Liang \\
School of Information Science \\ and Engineering, 
Shandong University \\
\texttt{liangdaojun@mail.sdu.edu.cn} 
\And
Haixia Zhang \\
School of Control Science \\ and Engineering, 
Shandong University \\
\texttt{haixia.zhang@sdu.edu.cn} \\
\And
Dongfeng Yuan \\
School of Qilu Transportation,\\ Shandong University \\
\texttt{dfyuan@sdu.edu.cn} 
}

\begin{document}

\maketitle

\begin{abstract}
  Traditional regression and prediction tasks often only provide deterministic point estimates. To estimate the distribution or uncertainty of the response variable, traditional methods either assume that the posterior distribution of samples follows a Gaussian process or require thousands of forward passes for sample generation. We propose a novel approach called DistPred for regression and forecasting tasks, which overcomes the limitations of existing methods while remaining simple and powerful. 
  Specifically, we transform proper scoring rules that measure the discrepancy between the predicted distribution and the target distribution into a differentiable discrete form and use it as a loss function to train the model end-to-end. This allows the model to sample numerous samples in a single forward pass to estimate the potential distribution of the response variable. 
  We have compared our method with several existing approaches on multiple datasets and achieved state-of-the-art performance. Additionally, our method significantly improves computational efficiency. For example, compared to state-of-the-art models, DistPred has a 180x faster inference speed. Experimental results can be reproduced through this \href{https://github.com/Anoise/DistPred}{Repository}.
\end{abstract}

\section{Introduction}

Traditional deterministic point estimates are no longer sufficient to meet the needs of AI safety and uncertainty quantification.
For example, we may want to obtain confidence intervals for predicted points to make important decisions, such as deciding whether to travel based on weather forecasts or how to invest based on stock predictions. Moreover, this is particularly important in high-security AI application areas such as autonomous driving, risk estimation, and decision-making.

\begin{figure*}[t]
  \centering
  \centerline{\includegraphics[width=0.95\textwidth]{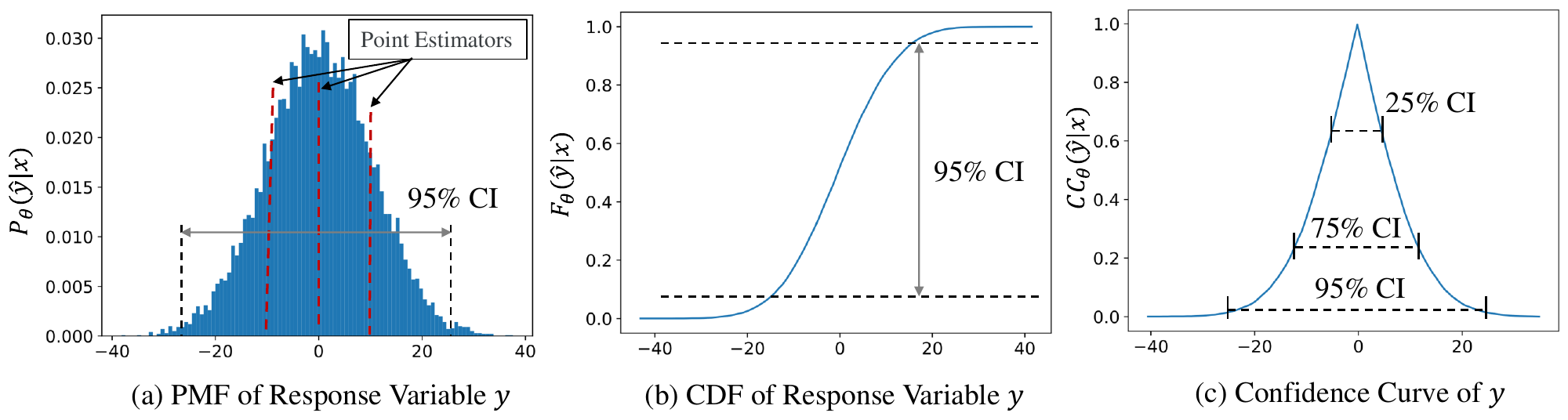}}
  \caption{DistPred can provide $K$ predicted values $\hat{y}$ of the response variable $y$ given the predictor variable $x$ in a single forward process, denoted as $\mathbb{E}(\hat{Y}|x)$, where $\hat{Y}$ represents a maximum likelihood sample of $y$. Based on this sampling, the probability mass function (PMF) $P_{\theta}(\hat{y}|x)$, cumulative distribution function (CDF) $F_{\theta}(\hat{y}|x)$, and confidence curve $CC_{\theta}(\hat{y}|x)$ for the response variable $y$ can be computed, thereby yielding comprehensive statistical insights into $y$. For instance, this includes confidence intervals (CI) at any desired level, as well as p-values.}
  \label{fig_distchart}
\end{figure*}

In this paper, we consider the underlying distribution behind predicting the response variable because it reflects the confidence intervals at all levels. For example, based on this distribution, we can calculate confidence intervals, coverage rate, and uncertainty quantification at any level, as shown in Fig. \ref{fig_distchart}. Currently, predicting the distribution of the response variable poses a challenge because at a specific moment, the response variable can only take on a single deterministic value. This point can be viewed as a `\textit{representative}' sample from its underlying distribution, but it fails to represent the overall state of the underlying distribution.

Currently, several methods are employed to predict the underlying distribution of the response variable in regression and forecasting tasks.
The most straightforward approach is to assume that the response variable follows a prior distribution and represents a specific statistic from that distribution. Specifically, several methods \citep{bishop1994mixture, greene2003econometric, salinas2020deepar, NEURIPS2020_NTK} transform distribution prediction and uncertainty quantification into predicting statistical variables such as mean and variance by assuming that the response variable follows a known continuous distribution. For instance,  mixture density networks \citep{bishop1994mixture} superimpose a specific distribution, typically Gaussian, weighted by designated parameters to fit the prior distribution.
Heteroscedasticity regression \citep{greene2003econometric} quantifies uncertainty by modeling the variability of residuals as a function of independent variables.
These methods only predict statistical variables, which reduces the inference cost, but strong assumptions often fail to capture the true data distribution, resulting in inferior performance. 

Conformal prediction offers an alternative approach for distribution prediction \citep{vovk2017nonparametric, vovk2018cross, romano2019conformalized, xu2021conformal, xu2023sequential}. 
The authors defined the random prediction system in \citep{vovk2017nonparametric, vovk2018cross} and proposed a nonparametric prediction distribution method based on conformal assumptions. 
The authors integrated conformal prediction with quantile regression in \citep{romano2019conformalized, xu2021conformal, xu2023sequential} to construct prediction intervals for the response variable by training multiple bootstrap estimators.
However, the application of conformal prediction to distribution prediction can be constrained by its reliance on the exchangeability assumption of residuals and the challenges in managing temporal dependencies (autocorrelation), potentially resulting in less reliable prediction intervals in non-i.i.d. data settings.

Uncertainty quantification is a method that indirectly reflects the potential distribution of the response variable, which can be classified into two primary categories: epistemic uncertainty and aleatoric uncertainty \cite{der2009aleatory}. Epistemic uncertainty refers to the uncertainty within the model itself. In contrast, aleatoric uncertainty pertains to the inherent randomness in the observations.
Quantitative analysis involves generating numerous samples (e.g., using MCMC) by perturbing the explanatory variables or models, thereby approximating the underlying distribution. For instance, bayesian neural networks (BNNs) \citep{wierstra2015weight, immer2021improving, daxberger2021laplace} simulate this uncertainty by assuming that their parameters follow a predefined distribution, thereby capturing the model's uncertainty given the data. Similarly, ensemble-based methods have been proposed to combine multiple deep models with random outputs to capture prediction uncertainty. MC Dropout \citep{gal2016dropout} shows that enabling dropout during each testing process yields results akin to model ensembling. Additionally, models based on GANs and diffusion have been introduced for conditional density estimation and prediction uncertainty quantification \citep{jointmatching, wganjointmatching, han2022card}. These models utilize noise during the generation or diffusion process to obtain different predicted values for estimating the uncertainty of the response variable.

The common characteristic of these methods mentioned above is the requirement of $K$ forward passes to sample $K$ representative samples. For example, Bayesian framework-based methods require $K$ learnable parameter samples to be inferred in order to obtain $K$ representative samples; ensemble methods require $K$ models to jointly infer; MC Dropout requires $K$ forward passes with random dropout activations; generative models require $K$ forward or diffusion processes. However, the excessive forward passes result in significant computational overhead and slow speed, a drawback that becomes increasingly apparent for AI applications with high real-time requirements.


To address this issue, we propose a novel method called DistPred, which is a distribution-free probabilistic inference method for regression and forecasting tasks. DistPred is a simple and powerful method that can estimate the distribution of the response variable in a single forward pass. Specifically, we contemplate employing all predictive quantiles to specify the potential cumulative density function (CDF) of the predictor variable, and we show that the full quantiles' prediction can be translated into calculating the minimum expected score of the response variable and the predictive ensemble variables. Based on this, we transform proper scoring rules that measure the discrepancy between the predicted distribution and the target distribution into a differentiable discrete form and use it as a loss function to train the model end-to-end. This allows the model to sample numerous samples in a single forward pass to estimate the potential distribution of the response variable. 
DistPred is orthogonal to other methods, enabling its combination with alternative approaches to enhance estimation performance. 
Further, we show that DistPred can provide comprehensive statistical insights into the response variable, including confidence intervals at any desired level, p-values, and other statistical information, as shown in Fig. \ref{fig_distchart}. 
Experimental results show that DistPred outperforms existing methods in terms of both accuracy and computational efficiency. Specifically, DistPred has a 180x faster inference speed than state-of-the-art models.

\begin{figure*}[ht]
  \centering
  \centerline{\includegraphics[width=\textwidth]{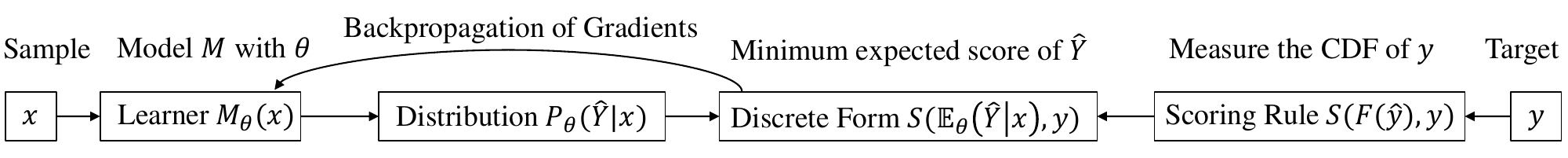}}
  \caption{The workflow of DistPred. An ensemble of predictive variables $\hat{Y}$ is inferred in a forward pass and $S(\mathbb{E}(\hat{Y}|x),y)$ is utilized to train the learner end-to-end.}
  \label{fig_workflow}
\end{figure*}

\section{Method}
\label{sec_method}

Assume that the dataset $D=\{x_i,y_i\}_{i=1}^N$ consists of $N$ sample-label pairs. The subscript $i$ will be omitted if it does not cause ambiguity in the context. Our objective is to utilize a machine learning model $M$ with parameters $\theta$ to predict the underlying distribution $P(y|x)$ from $D$, aiming to acquire comprehensive statistical insights such as obtaining confidence intervals (CI) and quantifying uncertainty at any desired level.

Direct prediction of distribution $P_\theta(\hat{y}|x)$ is not feasible because:

\begin{enumerate}
\item Without distributional assumptions, we cannot give a valid representation of the PDF or CDF for the response variable.
\item We have only a deterministic target point, without access to its distribution information, which limits our ability to guide the model's learning process.
\end{enumerate}

To address the aforementioned issues, we contemplate employing full predictive quantiles $\hat{q}_1, \hat{q}_2, \cdots, \hat{q}_K$ at levels $\alpha_1, \alpha_2, \cdots, \alpha_K$, $(K \rightarrow \infty)$, to specify the potential CDF $F_\theta(\hat{y})$ of the response variable $\hat{y}$.
This is because if we know the cumulative distribution function of a random variable, we can find any quantile by
\begin{equation}
  q_{\alpha} = \text{inf} \{ y \in \mathbb{R}: F(y) \ge \alpha \}.
\end{equation}
Conversely, if we have a complete set of quantiles, we can approximate or reconstruct the cumulative distribution function of the random variable
\begin{equation}
  F(y) = \text{sup} \{ \alpha \in [0, 1]: q_{\alpha} \le y \}.
\end{equation}
Full quantiles provide discrete `snapshots' of the distribution, while the CDF is a continuous, smooth version of these snapshots, offering a complete description of the cumulative probability from the minimum to the maximum value.

\subsection{Property of Full Quantiles}

We contemplate probabilistic predictions pertaining to a continuous quantity, manifested as full predictive quantiles $\hat{q}_1, \cdots, \hat{q}_K, (K \rightarrow \infty)$.
For $P\in \mathcal{P}$, let $\hat{q}_1, \cdots, \hat{q}_K$ denote the true $P$-quantiles at levels $\alpha_1, \cdots, \alpha_K \in (0,1)$.
Then, the expected score $S(q_1, \cdots, q_K; P)$ can be defined as
\begin{equation}
  S(\hat{q}_1, \cdots, \hat{q}_K; P) = \int S(\hat{q}_1, \cdots, \hat{q}_K; y)  \,\text{d}P(y). 
  \label{eq_quantile_scoring}
\end{equation}
The function $S$ in Eq. \ref{eq_quantile_scoring} satisfies the scoring rule, which offers a concise measure for assessing probabilistic forecasts by assigning numerical scores according to the forecast distribution and predicted outcomes \citep{gneiting2007strictly, jordan2017evaluating}.
Specifically, let $\Omega$ denote the set of possible values of the quantity of interest, and let $\mathcal{P}$ denote a convex class of probability distributions on $\Omega$, the scoring rule is a function
\begin{equation}
S: \Omega \times \mathcal{P} \rightarrow \mathbb{R} \cup \{\infty\} \label{eq_scoring_rule}
\end{equation}
that assigns numerical values to pairs of forecasts $P \in \mathcal{P}$ and observations $y \in \Omega$.
We identify probabilistic forecasts $P$ with the associated CDF $F$ or PDF $f$, and consider scoring rules to be negatively oriented, where a lower score signifies a more accurate forecast. A proper scoring rule is optimized when the forecast aligns with the true distribution of the observation, i.e., if 
\begin{equation}
  E_{y \sim Q} [S(Q, y)] \leq E_{y \sim Q} [S(P, y)] \label{eq_proper_scoring_rule}
\end{equation}
for all $P, Q \in \mathcal{P}$.  
A scoring rule is termed strictly proper when equality is achieved only when $P = Q$. Proper scoring rules are essential for comparative evaluation, especially in ranking forecasts. 

Based on this definition, we can get the scoring rule $S$ is proper when $ S(q_1, \cdots, q_K; P) \ge S( \hat{q}_1, \cdots, \hat{q}_K; P)$. 
Further, assuming that $s_k$ is a non-decreasing probability measure of $\mathcal{P}$, then the scoring rule
  \begin{equation}
    S( \hat{q}_1, \cdots, \hat{q}_K; P) = \sum_{k=1}^K \left( \alpha_k s_k(\hat{q}_k) + (s_k(y) - s_k(\hat{q_k})\mathbb{I}\{y\le \hat{q}_k\}  ) \right) 
    \label{eq_quantile_proper}
  \end{equation}
is proper for predicting the quantiles at levels $\alpha_1, \cdots, \alpha_K$ when $K \rightarrow \infty$ \citep{gneiting2007strictly, schervish2012characterization}. $\mathbb{I}\{y\le \hat{q}_k\}$ denotes the indicator function which is $1$ if $y \le \hat{q}_k$ and $0$ otherwise.


\subsection{Predicting CDF by Full Quantiles}

Eq. \ref{eq_quantile_proper} shows that full predictive quantiles are proper. 
Consequently, we can formulate scoring rules for the predictive distribution based on the scoring rules for the quantiles.
Specifically, let $S_\alpha$ denote a proper scoring rule for the quantile at level $\alpha$, then the scoring rule
\begin{equation}
  S(F,y) = \int_0^1 S_\alpha(F^{-1}(\alpha); y) \,\text{d}\alpha =  \int_{-\infty}^{\infty} S(F(\hat{y}), \mathbb{I}\{y \le \hat{y} \}) \,\text{d}\hat{y}  \label{eq_cdf_scoring}
\end{equation}
is proper. 
Here, we establish a relationship between the full quantiles of the response variable and its CDF. 
However, the CDF here is in continuous form and cannot be directly observed. Therefore, we need to convert it into a discrete form, as the number of quantiles $K$ is always finite in practice.
It is worth noting that Eq. \ref{eq_cdf_scoring} corresponds to the continuous ranked probability score (CRPS) in which $S$ is the quadratic or Brier score,  defined as
\begin{equation}
  C(F, y) = \int_{-\infty}^{\infty} (F(\hat{y}) - \mathbb{I}\{y \le \hat{y} \})^2 \,\text{d}\hat{y}. \label{eq_crps}
\end{equation}
When the first moment of $F$ is finite, the CRPS can be written as
\begin{equation}
  C(F, y) = \mathbb{E}_F[|\hat{Y} - y|] - \frac{1}{2} \mathbb{E}_{FF}[|\hat{Y} - \hat{Y}'|],
  \label{eq_crps2}
\end{equation}
where $\hat{Y}$ and $\hat{Y}'$ denote independent predictive ensemble variables with distribution $F$. 

Eq. \ref{eq_crps2} incentivizes forecasters to accurately report their perception of the true distribution in this scenario.
Therefore, it provides attractive measures and utility functions that can be tailored to a regression or forecast problem.
To estimate $\theta$, we need to measure the goodness-of-fit by the mean score
\begin{equation}
  C_N(\theta) = \frac{1}{N} \sum_{i=1}^N C(F_\theta(\hat{y}_i), y_i). \label{eq_mean_score}
\end{equation}
Let $\theta^*$ denote the true parameter value, then asymptotic arguments indicate that $\text{argmin}_\theta C_N(\theta) \rightarrow \theta^*$ as $N \rightarrow \infty$.
As shown in Fig. \ref{fig_workflow}, the workflow suggests a general approach to transforming proper scoring rules into loss functions for training models, which implicitly minimizes the discrepancy between predictive and true distributions.

\subsection{End-to-End Ensemble Inference}

\begin{wrapfigure}{r}{0.3\textwidth}
  \centering
  \vspace{-40pt}
  \includegraphics[width=0.3\textwidth]{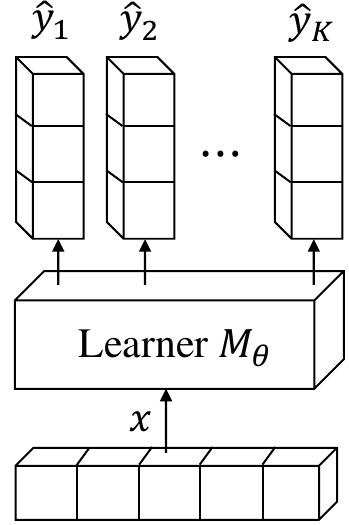}
  \caption{DistPred's architecture. A model $M$ with parameters $\theta$ takes one input variable $x$ and outputs an ensemble of $K$ response variable $\hat{y}_1, \cdots, \hat{y}_K$.}
  \label{fig_arch}
  \vspace{-24pt}
\end{wrapfigure}


Based on the analysis provided above, it is evident that predicting the full quantiles is equivalent to minimizing Eq. \ref{eq_crps2} w.r.t $\mathbb{E}(\hat{Y}|y)$. 
Hence, as illustrated in the architecture depicted in Fig. \ref{fig_arch}, we can develop a model $M$ with parameters $\theta$ that infers an ensemble of predictive variables $\hat{Y} = \{\hat{y}_1, \cdots, \hat{y}_K \}$ in a forward pass and utilize Eq. \ref{eq_mean_score} to train it end-to-end. 
This allows the model to sample numerous samples in a single forward pass to estimate the empirical CDF $\hat{F}$ by the predictive ensemble variables
\begin{equation}
  C(\hat{F},y) = \frac{1}{K}\sum_{k=1}^{K}|\hat{y}_k - y| - \frac{1}{2K^2}\sum_{k=1}^{K}\sum_{j=1}^{K}|\hat{y}_k - \hat{y}_{j}|.
  \label{eq_crps3}
\end{equation}
It's worth noting that Eq. \ref{eq_crps3} is a differentiable discrete form w.r.t $\hat{Y}$ and $\hat{Y'}$ that strictly satisfies proper scoring rules. 
However, implementations of Eq. \ref{eq_crps3} exhibit inefficiency due to their storage complexity of $\mathcal{O}(K^2)$.
This can be enhanced by using algebraically equivalent representations based on the generalized quantile function \citep{laio2007verification} and the sorted predictive ensemble variables $\overrightarrow{y_k}$,
\begin{equation}
  C(\hat{F},y) = \frac{1}{2K^2}\sum_{k=1}^{K}(\overrightarrow{y_k} - y)(k \mathbb{I}\{y \le \overrightarrow{y_k} \} - k + \frac{1}{2}).
  \label{eq_crps4}
\end{equation}
Since a sorting operation is involved, the storage complexity of Eq. \ref{eq_crps4} is $\mathcal{O}(KlogK)$.
However, since Eq. \ref{eq_crps4} renders the objective $C(\hat{F},y)$ non-differentiable, we employ probability-weighted moment estimation form to approximate it, as
\begin{equation}
  C(\hat{F},y) = \frac{1}{K}\sum_{k=1}^{K}|\hat{y}_k - y| + \frac{1}{K}\sum_{k=1}^{K} \hat{y}_k - \frac{2}{K(K-1)}\sum_{k=1}^{K} \hat{y}_k (k-1) .
  \label{eq_crps_effective}
\end{equation}
Actually, according to the definition of L-Moments, we can prove that the last term of Eq. \ref{eq_crps3} can be decomposed into the 1-th and 0-st order L-moments of Eq. \ref{eq_crps_effective}, i.e., Eq. \ref{eq_crps_effective} is an unbiased estimate of Eq. \ref{eq_crps3} when it has a finite first L-moment.
To conserve memory, we suggest utilizing Eq. \ref{eq_crps_effective}, which has a storage complexity of $\mathcal{O}(K)$, as the loss function, since predicting ensemble variables in long-term forecasting tasks may lead to out-of-memory issues on GPUs.

\subsection{Incorporate Alternative Methodologies}

DistPred is orthogonal to other methods, enabling its combination with alternative approaches to enhance estimation performance. 
Here, with a focus on computational efficiency and memory conservation, we opt to integrate MC Dropout with DistPred, thereby denoting the amalgamation as \textbf{DistPred+MCD}. 
In our experiments, we observed that using DistPred+MCD can further enhance uncertainty quantification performance, albeit with a marginal increase in computational effort.

\begin{figure*}[ht]
  \centering
  \centerline{\includegraphics[width=\textwidth]{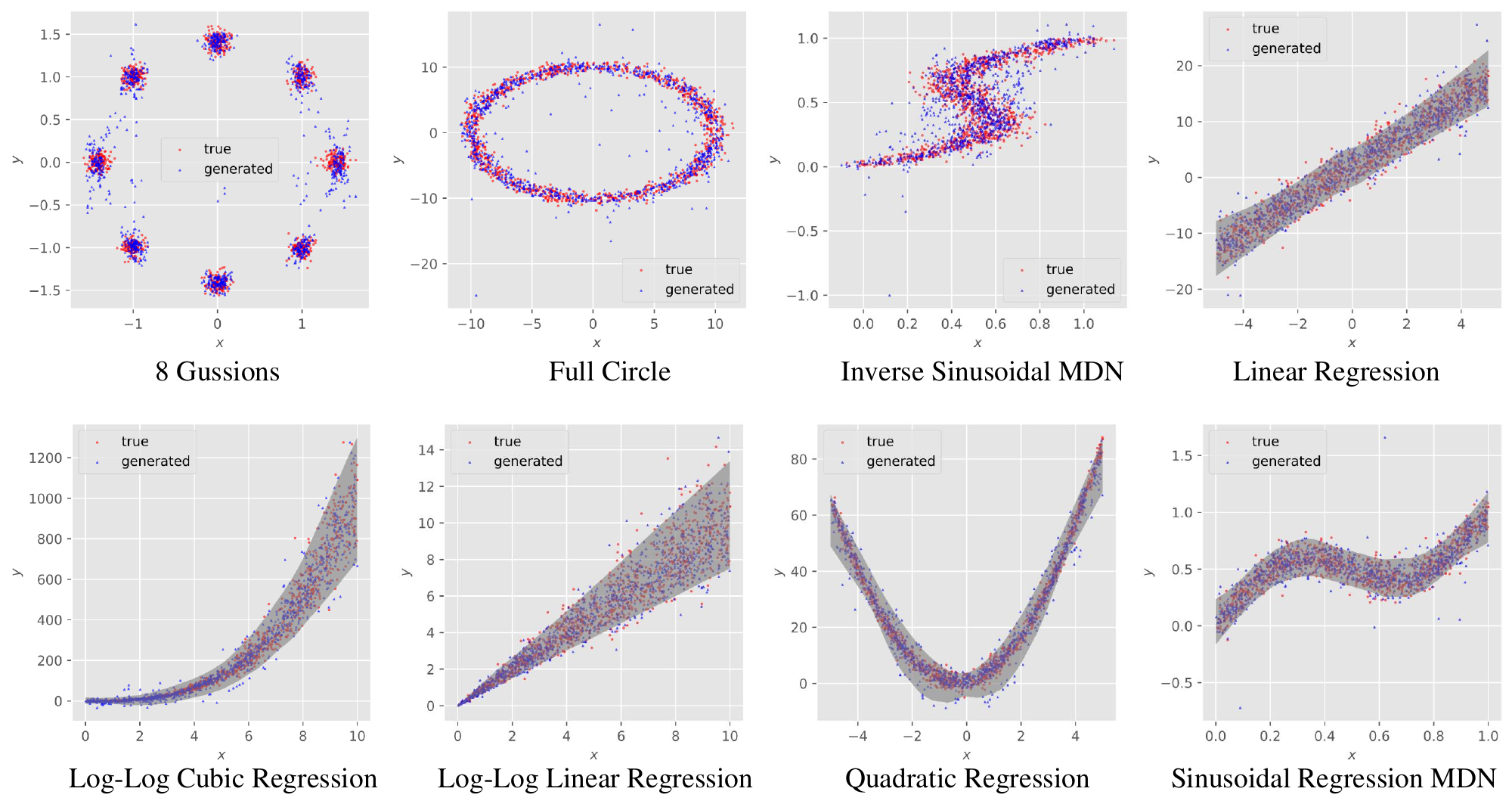}}
  \caption{Scatter plot of DistPred's regression results on $8$ toy examples.}
  \label{fig_toy_exp}
\end{figure*}

\section{Experiments}

In this paper, our focus centers on regression (proposed by \citet{hernandez2015probabilistic}) and prediction (proposed by \citet{Zhou2021Informer}) tasks, where we validate the application of the proposed DistPred method to these specific endeavors.

\subsection{PICP and QICE Metrics}

The metric utilized in BNNs to assess uncertainty estimates, namely the negative log-likelihood (NLL), is computed based on Gaussian prior. This assumption implies that they consider both the conditional distribution $p(y|x=x')$ for all $x'$ are Gaussian.
However, this assumption is very difficult to verify for real-world datasets. 
We follow \citep{han2022card} and use the following two metrics, both of which are designed to empirically evaluate the degree of similarity between learned and true conditional distributions:
\begin{itemize}
  \item \textbf{PICP} (Prediction Interval Coverage Probability) \citep{yao2019quality} is a metric that measures the proportion of true labels that fall within the prediction interval.
  \item \textbf{QICE} (Quantile Interval Calibration Error) \citep{han2022card} is a metric that measures the average difference between the predicted and true quantiles at a given level $\alpha$.
\end{itemize}

The PICP is calculated as
\begin{equation}
  PICP := \frac{1}{N} \sum_{n=1}^{N} \mathbb{I}\{\hat{y}_n \ge q_{\alpha/2} \} \cdot \mathbb{I}\{\hat{y}_n \le q_{1-\alpha/2} \},
\end{equation}
where $q_{\alpha/2}$ and $q_{1-\alpha/2}$ represent the low and high percentiles, respectively, that we have selected for the predicted $\hat{y}$ outputs given the same $x$ input.
This metric evaluates the proportion of accurate observations that lie within the percentile range of the generated $\hat{y}$ samples corresponding to each $x$ input.
Within this study, we opt for the $2.5$th and $97.5$th percentiles, signifying that an optimal PICP value for the model should ideally reach 95\%.

However, a caveat of the PICP metric becomes apparent in the measurement of distribution differences. 
Drawing from this reasoning, \citet{han2022card} introduces a novel empirical metric called QICE. This metric can be perceived as an enhanced version of PICP, offering finer granularity and addressing the issue of uncovered quantile ranges.
To calculate QICE, the initial step involves generating an adequate number of samples for each $\hat{y}$ value. These samples are then divided into $M$ bins of approximately equal sizes. Subsequently, the quantile values are determined at each boundary within these bins.
The definition of QICE entails computing the mean absolute error (MAE) between the proportion of true data encompassed by each quantile interval and the optimal proportion, which is 1/M for all intervals:
\begin{align}
  QICE := & \frac{1}{M}\sum_{m=1}^{M} | r_m - \frac{1}{M} |, \\
  \text{where} \ \ r_m = & \frac{1}{N} \sum_{n=1}^{N} \mathbb{I}\{\hat{y}_n \ge q_{\alpha/2} \} \cdot \mathbb{I}\{\hat{y}_n \le q_{1-\alpha/2} \}. \notag
\end{align}
In this paper, we followed \citep{han2022card} and set $M=10$ for all experiments. 

\subsection{Toy Examples}
To demonstrate the effectiveness of DistPred, we initially conducted experiments on $8$ toy examples as done in \citep{han2022card}. These examples are specifically crafted with distinct statistical characteristics in their data generating functions: some have a uni-modal symmetric distribution for their error term (linear regression, quadratic regression, sinusoidal regression), while others exhibit heteroscedasticity (log-log linear regression, log-log cubic regression) or multi-modality (inverse sinusoidal regression, $8$ Gaussians, full circle).

The research demonstrates that a trained DistPred model has the capability to produce samples that closely resemble the true response variable for novel covariates. Additionally, it can quantitatively match the true distribution based on certain summary statistics. The study visualizes scatter plots comparing real and generated data for all eight tasks in Fig. \ref{fig_toy_exp}. In cases where the tasks involve unimodal conditional distributions, the interest region fills the region between the $2.5$th and $97.5$th percentiles of the generated $\hat{y}$ values.

We note that within every task, the generated samples seamlessly integrate with the authentic test instances, indicating the potential of DistPred to reconstruct the inherent data generation process. 
This experiment visually demonstrates that DistPred effectively reconstructs the sample potential distribution of the target response variable. This indicates that the advantages of DistPred mentioned earlier can be fully harnessed in distribution prediction.

\begin{figure}
  \centering
  \centerline{\includegraphics[width=0.65\textwidth]{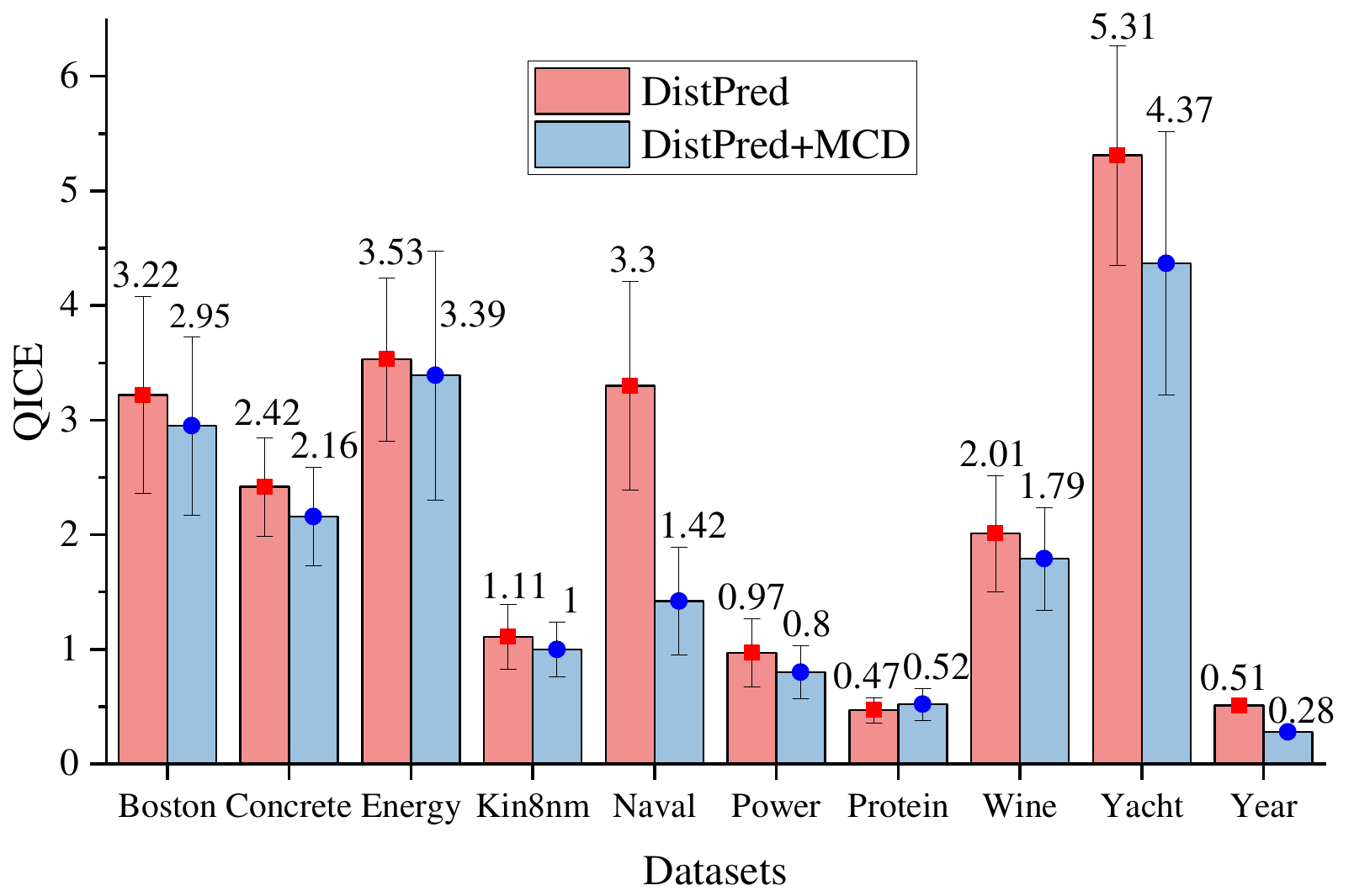}}
  \caption{DistPred and DistPred+MCD on UCI datasets.}
  \label{fig_dist_mcd}
\end{figure}

\begin{figure*}[!ht]
  \centering
  \centerline{\includegraphics[width=0.9\textwidth]{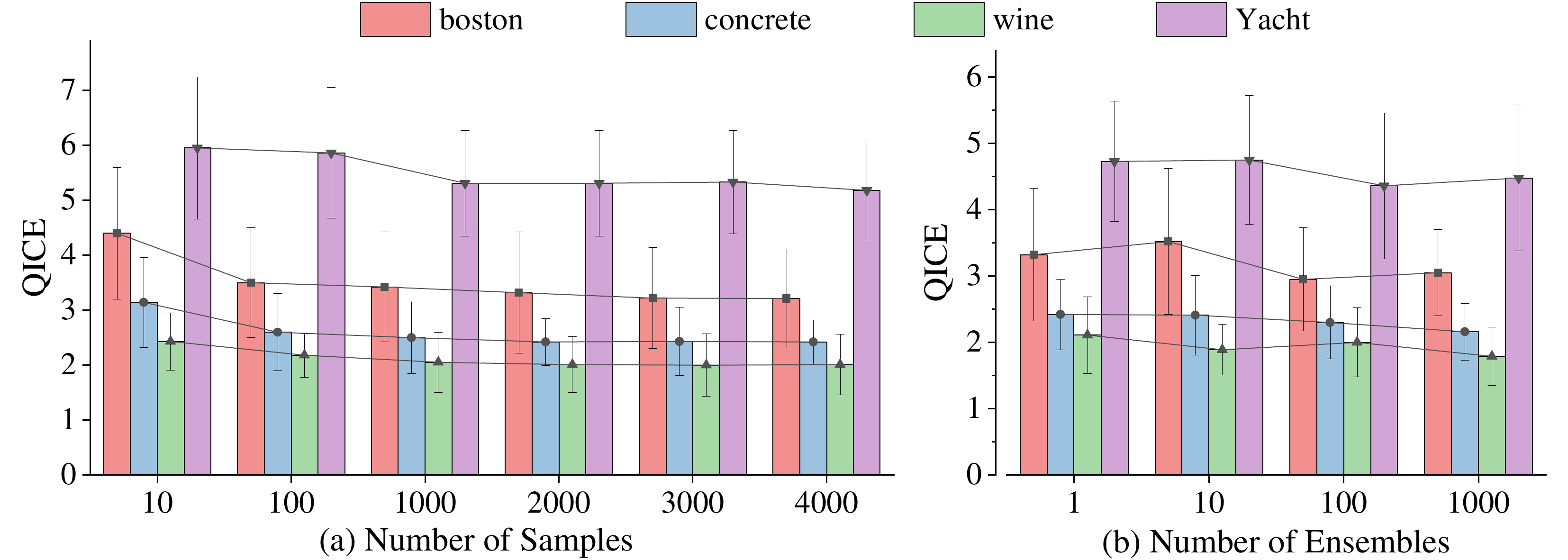}}
  \caption{Ablation studies of the number of samples in DistPred (a) and the number of ensembles in DistPred+MCD (b).}
  \label{fig_ensemble}
\end{figure*}

\subsection{UCI Regression Tasks}

\begin{table}
  \centering
  \caption{Comparison of model training and inference times (minutes) on UCI regression datasets when $K=1000$.}
  \resizebox{0.85\textwidth}{!}
  {
  
  \begin{tabular}{cccccc}
    \toprule
    Models    & DistPred          & DistPred+MCD       &  PBP                & MC Dropout          & CARD         \\
    \midrule
    Training  & 0.035 $\pm$ 6E-3  & 0.035  $\pm$ 6E-3  &  0.04 $\pm$ 8E-3    & 0.031 $\pm$ 0.01    & 8.14$\pm$  0.05  \\
    \midrule
    Inference & 0.026 $\pm$ 4E-3  & 0.027  $\pm$ 5E-3  &  5.23 $\pm$ 0.10   & 4.62 $\pm$  0.06    & 8.31 $\pm$ 0.17 \\ 
    \bottomrule
  \end{tabular}
  }
  \label{tb_speed}
\end{table}

\begin{figure*}
  \centering
  \centerline{\includegraphics[width=\textwidth]{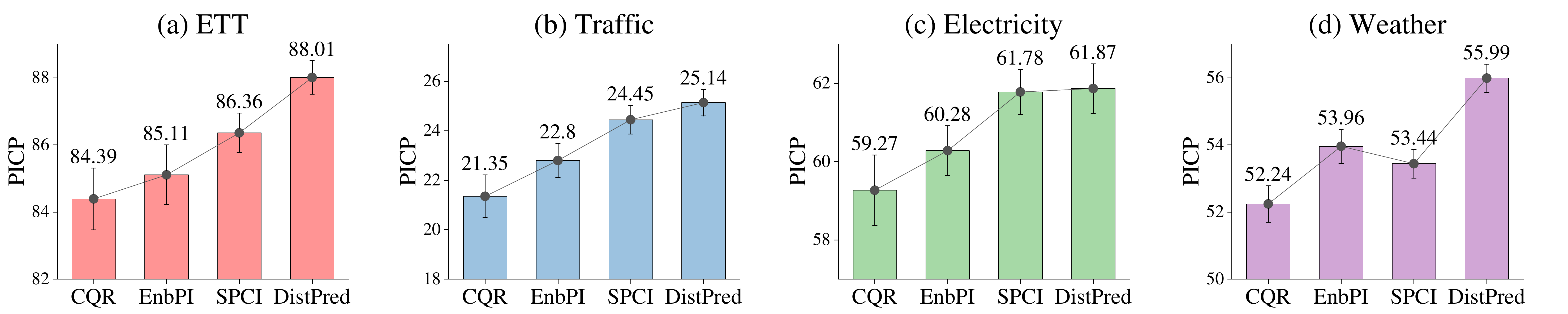}}
  \caption{Comparison DistPred (ours) with conformal prediction methods. All results are averaged across all prediction lengths.}
  \label{fig_vs_cp}
\end{figure*}

For experiments conducted on real-world datasets, we utilize the same $10$ UCI regression benchmark datasets \citep{asuncion2007uci} and follow the experimental protocol introduced by \citet{hernandez2015probabilistic}, which has also been followed by \citet{gal2016dropout} and \citet{lakshminarayanan2017simple}, as well as by \citet{han2022card}.
The dataset information can be found in Table \ref{tb_reg_uci_dim} located in Appendix \ref{app_dataset}.

\begin{table*}
    \begin{center}
      \caption{\label{tb_uci_qice}QICE $\downarrow$ (in $\%$) of UCI regression tasks.}\vspace{1mm}
      \resizebox{0.85\textwidth}{!}{
      \begin{tabular}{@{}lccccccc@{}}
      \toprule[1.5pt]
        Dataset        & PBP                 & MC Dropout          & Deep Ensembles      & GCDS                & CARD   & DistPred  & DistPred-MCD           \\ \midrule
      Boston   & $3.50\pm 0.88$      & $3.82\pm 0.82$ & $\underline{3.37\pm 0.00}$     & $11.73\pm 1.05$ & $3.45\pm 0.83$  & $\underline{3.22 \pm 0.86}$  & $\bm{2.95 \pm 0.78}$  \\
      Concrete & $2.52\pm 0.60$      & $4.17\pm 1.06$ & $2.68\pm 0.64$     & $10.49\pm 1.01$ & $\underline{2.30\pm 0.66}$ &  $\underline{2.42 \pm  0.43}$  & $\bm{2.16 \pm 0.43}$ \\
      Energy   & $6.54\pm 0.90$      & $5.22\pm 1.02$ & $\underline{3.62\pm 0.58}$ & $7.41\pm 2.19$ & $4.91\pm 0.94$     & $\underline{3.73 \pm 0.71}$  &  $\bm{3.39 \pm 1.09}$  \\
      Kin8nm   & $1.31\pm 0.25$      & $1.50\pm 0.32$ & $1.17\pm 0.22$     & $7.73\pm 0.80$ & $\bm{0.92\pm 0.25}$ & $ \underline{1.11 \pm 0.28}  $   & $\underline{1.00 \pm 0.24}$  \\
      Naval    & $4.06\pm 1.25$ & $12.50\pm 1.95$ & $6.64\pm 0.60$     & $5.76\pm 2.25$ & $\bm{0.80\pm 0.21}$     & $ \underline{3.30 \pm 0.91} $   & $\underline{1.42 \pm 0.47}$  \\
      Power    & $\underline{0.82\pm 0.19}$  & $1.32\pm 0.37$ & $1.09\pm 0.26$     & $1.77\pm 0.33$ & $\underline{0.92\pm 0.21}$     & $ 0.97 \pm  0.30$   & $ \bm{0.80 \pm 0.23} $  \\
      Protein  & $1.69\pm 0.09$      & $2.82\pm 0.41$ & $2.17\pm 0.16$     & $2.33\pm 0.18$ & $\underline{0.71\pm 0.11}$  & $\bm{ 0.47 \pm 0.11} $   & $\underline{0.52 \pm 0.14}$  \\
      Wine     & $\underline{2.22\pm 0.64}$  & $2.79\pm 0.56$ & $2.37\pm 0.63$     & $3.13\pm 0.79$ & $3.39\pm 0.69$     & $\underline{2.01 \pm 0.51}$   & $ \bm{1.79 \pm 0.45} $  \\
      Yacht    & $6.93\pm 1.74$      & $10.33\pm 1.34$ & $7.22\pm 1.41$     & $\underline{5.01\pm 1.02}$ & $8.03\pm 1.17$     & $\underline{5.31 \pm  0.96}$   & $\bm{4.37 \pm  1.15}$  \\
      Year     & $2.96\pm$ NA           & $2.43\pm$ NA            & $2.56\pm$ NA          & $1.61\pm$ NA      & $\underline{0.53\pm \text{NA} }$  & $\underline{0.58 \pm \text{NA}}$  & $\bm{0.28 \pm \text{NA}}$      \\ \midrule
      \# Top 1   & $0$      & $0$            & $0$      & $0$      &  $2 $  & $1$  & $\bm{7}$    \\ \midrule
      \# Top 2 & $2$      & $0$            & $2$      & $1$      & $ 6$  & $9$  & $10$    \\ \bottomrule[1.5pt]
      \end{tabular}
      }
    \end{center}
  \end{table*}

We compare DistPred with other state-of-the-art methods, including PBP \citep{hernandez2015probabilistic}, MC Dropout \citep{gal2016dropout}, DeepEnsemble \citep{lakshminarayanan2017simple}, and another deep generative model that estimates a conditional distribution sampler, GCDS \citep{jointmatching}, as well as the  diffusion model, CARD \citep{han2022card}.
The multiple train-test splits are applied with a 90\%/10\% ratio, following the same methodology as \citet{hernandez2015probabilistic} and \citet{han2022card} (20 folds for all datasets except 5 for Protein and 1 for Year). The reported metrics are presented as the mean and standard deviation across all splits.
As pointed out by \citet{han2022card}, we compare the QICE of different methods on various UCI datasets.
Additional information regarding the experimental setup for these models is available in Appendix \ref{app_implementation}. The experimental results, along with corresponding metrics, are presented in Table \ref{tb_uci_qice}. 
The frequency with which each model achieves the best corresponding metric is reported in the penultimate row, while the frequency with which it achieves the top two positions is reported in the last row.

The results demonstrate that DistPred outperforms existing methods, often by a considerable margin.
It is worth noting that these impressive results are achieved in a single forward pass of the DistPred method.
Crucially, as shown in Fig. \ref{fig_dist_mcd}, the performance of uncertain quantization can be further enhanced by leveraging DistPred+MCD, a hybrid approach that combines DistPred and MC Dropout.

The implementation of DistPred on UCI regression tasks follows a straightforward approach: We employ a basic multilayer perceptron (MLP) as the foundational framework, complemented by Eq. \ref{eq_crps_effective} serving as the loss function for end-to-end training.
Due to the fact that DistPred necessitates solely a single forward inference, its inference speed is notably rapid.
Table \ref{tb_speed} presents a comparison of the training and inference speeds of mainstream models. 
It should be noted that, for a fair comparison, the implementations of various models are constructed on the same backbone and utilize the same equipment.
It is evident that DistPred is approximately achieves at least 180x faster inference speed compared to state-of-the-art models.
The inference speed of DistPred is slower than its training speed because it involves calculating distribution statistical metrics like QICE and PICP.

\subsection{Ablation Study of The Number of Samples and Ensembles}

We investigate the influence of the number of samples generated by DistPred, as well as the number of ensembles of DistPred+MCD, on their respective performances.
In Fig. \ref{fig_ensemble}(a), we increase the number of samples of DistPred from 10 to 4000 to observe the changes in its QICE.
In Fig. \ref{fig_ensemble}(b), we increase the number of ensembles of DistPred+MCD from 1 (DistPred) to 1000 to observe the changes in its QICE.
It can be found that with an increase in the number of output samples and ensembles, the model's performance shows a gradual improvement.

\begin{table*} 
    \centering
    \caption{Multivariate time series forecasting results on six benchmark datasets.}
    \label{tb2}
    \resizebox{\textwidth}{!}
    {
        \begin{tabular}{cc|ccccc| cc| cc| cc| cc| cc| cc| cc| cc}
        \toprule
        \multicolumn{2}{c}{Model}  & \multicolumn{5}{c}{DistPred} & \multicolumn{2}{c}{iTransformer} & \multicolumn{2}{c}{PatchTST} & \multicolumn{2}{c}{SCINet} & \multicolumn{2}{c}{TimesNet} & \multicolumn{2}{c}{DLinear} & \multicolumn{2}{c}{FEDformer} & \multicolumn{2}{c}{Autoformer} & \multicolumn{2}{c}{Informer} \\ \toprule
        \multicolumn{2}{c|}{Input Length}  & \multicolumn{5}{c|}{96} & \multicolumn{2}{c|}{96} & \multicolumn{2}{c|}{336} &  \multicolumn{2}{c|}{168} & \multicolumn{2}{c|}{96} & \multicolumn{2}{c|}{336} & \multicolumn{2}{c|}{96} & \multicolumn{2}{c|}{96} & \multicolumn{2}{c}{96} \\ \toprule
                                  & Output          & CRPS    &  QICE   & PICP       & MSE       & MAE & MSE              & MAE              & MSE             & MAE                     & MSE            & MAE           & MSE            & MAE            & MSE            & MAE            & MSE               & MAE          & MSE             & MAE             & MSE           & MAE             \\ \toprule
        \multirow{5}{*}{\rotatebox{90}{ETT}} 
          & 96   & 0.248   & 9.53  & 48.50      & {\bf 0.288}     & {\bf 0.334}   & \underline{0.297}            & \underline{0.349}            & 0.302           & 0.348                 & 0.707          & 0.621         & 0.340          & 0.374          & 0.333          & 0.387          & 0.358             & 0.397        & 0.346           & 0.388           & 3.755         & 1.525           \\
          & 192  & 0.277    & 9.22       & 51.06  & {\bf0.348}     & {\bf 0.371}     & \underline{0.380}            & \underline{0.400}            & 0.388           & 0.400                  & 0.860          & 0.689         & 0.402          & 0.414          & 0.477          & 0.476          & 0.429             & 0.439        & 0.456           & 0.452           & 5.602         & 1.931           \\
          & 336  & 0.298    & 8.91       & 53.28  & {\bf0.392}     & {\bf0.402}     & \underline{0.428}            & \underline{0.432}            & 0.426           & 0.433                   & 1.000          & 0.744         & 0.452          & 0.452          & 0.594          & 0.541          & 0.496             & 0.487        & 0.482           & 0.486           & 4.721         & 1.835        \\
          & 720  & 0.322    & 8.23   & 57.11 & 0.437  & {\bf 0.436}     & {\bf0.427}            & \underline{0.445}            & \underline{0.431}           & 0.446                   & 1.249          & 0.838         & 0.462          & 0.468          & 0.831          & 0.657          & 0.463             & 0.474        & 0.515           & 0.511           & 3.647         & 1.625        \\ \cline{2-23}
          & Avg  & 0.286   & 8.97   &  52.45   & {\bf0.366}     & {\bf0.386}      & \underline{0.383}            & \underline{0.407}            & 0.387           & 0.407               & 0.954          & 0.723         & 0.414          & 0.427          & 0.559          & 0.515          & 0.437             & 0.449        & 0.450           & 0.459           & 4.431         & 1.729           \\ \toprule
       
       \multirow{5}{*}{\rotatebox{90}{Traffic}}  
          & 96   & 0.193   & 12.75    & 34.85    & {\bf0.391}     & {\bf0.251}  & \underline{0.395}            & \underline{0.268}            & 0.544           & 0.359                   & 0.788          & 0.499         & 0.593          & 0.321          & 0.650          & 0.396          & 0.587             & 0.366        & 0.613           & 0.388           & 0.719         & 0.391           \\
          & 192 & 0.197    & 12.81   & 34.06   & {\bf0.416}     & {\bf0.269}   & \underline{0.417}            & \underline{0.276}            & 0.540           & 0.354          & 0.530                  & 0.505         & 0.617          & 0.336          & 0.598          & 0.370          & 0.604             & 0.373        & 0.616           & 0.382           & 0.696         & 0.379           \\
          & 336  & 0.202   & 12.90     & 34.03  & {\bf0.427}     & {\bf0.275}  & \underline{0.433}            & \underline{0.283}            & 0.551           & 0.358          & 0.558                  & 0.508         & 0.629          & 0.336          & 0.605          & 0.373          & 0.621             & 0.383        & 0.622           & 0.337           & 0.777         & 0.420           \\
          & 720 & 0.218    & -   & -   & {\bf0.467}     & {\bf0.297}   & {\bf0.467}            & 0.302            & 0.586           & 0.375                  & 0.841          & 0.523         & 0.640          & 0.350          & 0.645          & 0.394          & 0.626             & 0.382        & 0.660           & 0.408           & 0.864         & 0.472           \\ \cline{2-23}
          & Avg  & 0.203  & -    & - & {\bf0.425}     & {\bf0.273}    & \underline{0.428}            & \underline{0.282}            & 0.555           & 0.362                  & 0.804          & 0.509         & 0.620          & 0.336          & 0.625          & 0.383          & 0.610             & 0.376        & 0.628           & 0.379           & 0.764         & 0.416           \\ \toprule
        
        \multirow{5}{*}{\rotatebox{90}{Electricity}} 
          & 96  & 0.167   & 12.16     & 34.11   & {\bf0.138}     & {\bf0.231}   & \underline{0.148}            & \underline{0.240}            & 0.195           & 0.285                   & 0.247          & 0.345         & 0.168          & 0.272          & 0.197          & 0.282          & 0.193             & 0.308        & 0.201           & 0.317           & 0.274         & 0.368           \\
          & 192 & 0.178   & 11.62    & 38.85 & {\bf 0.155}     & {\bf 0.246} & \underline{0.162}            & \underline{0.253}            & 0.199           & 0.289                  & 0.257          & 0.355         & 0.184          & 0.289          & 0.196          & 0.285          & 0.201             & 0.315        & 0.222           & 0.334           & 0.296         & 0.386           \\
          & 336 & 0.188   & 11.14   & 41.27  & {\bf 0.169}     & {\bf 0.264}    & \underline{0.178}            & \underline{0.269}            & 0.215           & 0.305        & 0.269          & 0.369         & 0.198          & 0.300          & 0.209          & 0.301          & 0.214             & 0.329        & 0.231           & 0.338           & 0.300         & 0.394           \\
          & 720 & 0.209    & 10.79   & 42.83   & {\bf0.207}     & {\bf0.298}   & \underline{0.225}            &\underline{ 0.317}            & 0.256           & 0.337        & 0.299          & 0.390         & 0.220          & 0.320          & 0.245          & 0.333          & 0.246             & 0.355        & 0.254           & 0.361           & 0.373         & 0.439           \\ \cline{2-23}
          & Avg  & 0.186   & 11.45   & 39.27    & {\bf0.167}     & {\bf0.260}   & \underline{0.178}            & \underline{0.270}            & 0.216           & 0.304          & 0.268          & 0.365         & 0.192          & 0.295          & 0.212          & 0.300          & 0.214             & 0.327        & 0.227           & 0.338           & 0.311         & 0.397           \\ \toprule
        
        \multirow{5}{*}{\rotatebox{90}{Weather}}     
          & 96 & 0.145   & 12.05    & 29.96   & {\bf0.152}     & {\bf0.192}    & \underline{0.174}            & \underline{0.214}            & 0.177           & 0.218           & 0.221          & 0.306         & 0.172          & 0.220          & 0.196          & 0.255          & 0.217             & 0.296        & 0.266           & 0.336           & 0.300         & 0.384           \\
          & 192 & 0.185  & 10.607   & 39.85   & {\bf0.210}     & {\bf0.247}     & \underline{0.221}            & \underline{0.254}            & 0.225           & 0.259         & 0.261          & 0.340         & 0.219          & 0.261          & 0.237          & 0.296          & 0.276             & 0.336        & 0.307           & 0.367           & 0.598         & 0.544           \\
          & 336 & 0.216   & 9.857   & 44.84 & {\bf0.263}     & {\bf0.286}      & \underline{0.278}            & \underline{0.296}            & 0.278           & 0.297         & 0.309          & 0.378         & 0.280          & 0.306          & 0.283          & 0.335          & 0.339             & 0.380        & 0.359           & 0.395           & 0.578         & 0.523           \\
          & 720  & 0.263  & 9.31  & 47.83  & 0.362     & \underline{0.349}     & 0.358            & \underline{0.349}            & \underline{0.354}           & {\bf0.348}           & 0.377          & 0.427         & 0.365          & 0.359          & {\bf0.345}          & 0.381          & 0.403             & 0.428        & 0.419           & 0.428           & 1.059         & 0.741           \\ \cline{2-23}
          & Avg  & 0.202   & 10.46   & 40.62  & {\bf0.247}      & {\bf0.269}      & \underline{0.258}            & \underline{0.279}            & 0.259           & 0.281        & 0.292          & 0.363         & 0.259          & 0.287          & 0.265          & 0.317          & 0.309             & 0.360        & 0.338           & 0.382           & 0.634         & 0.548           \\ \toprule
        
        \multirow{5}{*}{\rotatebox{90}{Solar}} 
          & 96  & 0.152   & 6.36     & 65.75  & \underline{0.205}     & {\bf0.227}     & {\bf0.203}            & \underline{0.237}            & 0.234           & 0.286         & 0.237          & 0.344         & 0.250          & 0.292          & 0.290          & 0.378          & 0.242             & 0.342        & 0.884           & 0.711           & 0.236         & 0.259           \\
          & 192 & 0.165   & 8.39     & 53.34   & 0.236      & {\bf0.251}   & \underline{0.233}            & \underline{0.261}            & 0.267           & 0.310          & 0.280          & 0.380         & 0.296          & 0.318          & 0.320          & 0.398          & 0.285             & 0.380        & 0.834           & 0.692           & {\bf0.217}         & 0.269           \\
          & 336 & 0.171   & 7.82   & 55.37   & \underline{0.256}     & {\bf0.264}    & {\bf0.248}            & \underline{0.273}            & 0.290           & 0.315         & 0.304          & 0.389         & 0.319          & 0.330          & 0.353          & 0.415          & 0.282             & 0.376        & 0.941           & 0.723           & 0.249         & 0.283           \\
          & 720 & 0.187   & 8.79   & 55.15    & 0.273     & \underline{0.278}   & \underline{0.249}            & {\bf0.275}            & 0.289           & 0.317         & 0.308          & 0.388         & 0.338          & 0.337          & 0.356          & 0.413          & 0.357             & 0.427        & 0.882           & 0.717           & {\bf0.241}         & 0.317           \\ \cline{2-23}
          & Avg & 0.169   & 7.84    & 57.40 & 0.243     & {\bf0.255}      & {\bf0.233}            & \underline{0.262}            & 0.270           & 0.307          & 0.282          & 0.375         & 0.301          & 0.319          & 0.330          & 0.401          & 0.291             & 0.381        & 0.885           & 0.711           & \underline{0.235}         & 0.280           \\ \toprule
        
        \multirow{5}{*}{\rotatebox{90}{PEMS}} 
          & 12  & 0.120   & 9.01   & 52.99  & {\bf0.064}     & {\bf0.166}   & 0.071            & 0.174            & 0.099           & 0.216                  & \underline{0.066}          & \underline{0.172}         & 0.085          & 0.192          & 0.122          & 0.243          & 0.126             & 0.251        & 0.272           & 0.385           & 0.126         & 0.233           \\
          & 24  & 0.141  & 9.05   & 52.01   & \underline{0.087}     & {\bf0.192}    & 0.093            & 0.201            & 0.142           & 0.259          & {\bf0.085}          & \underline{0.198}         & 0.118          & 0.223          & 0.201          & 0.317          & 0.149             & 0.275        & 0.334           & 0.440           & 0.139         & 0.250           \\
          & 36  & 0.170  & 8.41   & 57.39  & {\bf0.124}     & {\bf0.233}    & \underline{0.125}            & \underline{0.236}            & 0.211           & 0.319         & 0.127          & 0.238         & 0.155          & 0.260          & 0.333          & 0.425          & 0.227             & 0.348        & 1.032           & 0.782           & 0.186         & 0.289           \\
          & 48  & 0.181   & 9.37   & 49.55  & {\bf0.140}     &{\bf0.245}     & \underline{0.160}            & \underline{0.270}            & 0.269           & 0.370          & 0.178          & 0.287         & 0.228          & 0.317          & 0.457          & 0.515          & 0.348             & 0.434        & 1.031           & 0.796           & 0.233         & 0.323           \\  \cline{2-23}
          & Avg & 0.153  & 8.96   & 52.99 & {\bf0.104}     & {\bf0.209}      & \underline{0.113}            & \underline{0.221}            & 0.180           & 0.291         & 0.114          & 0.224         & 0.147          & 0.248          & 0.278          & 0.375          & 0.213             & 0.327        & 0.667           & 0.601           & 0.171         & 0.274           \\ \toprule
        
        \multicolumn{2}{c|}{\# Top 1} & -   & -  & - & {\bf22}   & {\bf28}      & \underline{5}   & \underline{1}      & 0    & 1    & 1  & 0   & 0               & 0         & 1  & 0    & 0       & 0   & 0     & 0     & 2      & 0          \\  \bottomrule
        \end{tabular}
    }
  \end{table*}





\subsection{Time Series Distribution Forecasting}

We extend time series forecasting \citep{Zhou2021Informer,wu2021autoformer,zhou2022fedformer, liu2023itransformer} from point estimation to the task of distribution prediction to infer about more statistical information about a certain moment.

\textbf{Baselines}:
We employ recent 10 SOTA methods for comparisons, including iTransformer \cite{liu2023itransformer}, PatchTST \cite{nie2022time}, SCINet \cite{liu2022scinet}, TimesNet \cite{wu2022timesnet}, DLinear \cite{zeng2023transformers}, FEDformer \cite{zhou2022fedformer}, Autoformer \cite{wu2021autoformer}, Informer \cite{Zhou2021Informer}, LogTrans \cite{li2019enhancing} and Reformer \cite{Kitaev2020Reformer}.
DistPred employs the same network architecture as iTransformer.
We use the same experimental setup as \citep{Zhou2021Informer} and \citep{liu2022scinet} and follow the same experimental protocol as \citep{Zhou2021Informer}.
Univariate results can be found in Appendix \ref{app_univariate}.

\textbf{Datasets and Setting}:
The detailed information pertaining to the datasets can be located in Appendix \ref{app_dataset}.
The models \cite{liang2024minusformer} used in the experiments are evaluated over a wide range of prediction lengths to compare performance on different future horizons: 96, 192, 336, and 720. 
The experimental settings are the same for both multivariate and univariate tasks.
We use the average of the MSE and MAE ($\frac{MSE+MAE}{2}$) to evaluate the overall performance of the model.
It is noteworthy that DistPred provides an ensemble $\hat{Y}$ of response variable. Consequently, we employ the mean value of $\hat{Y}$ as the point estimate at that moment. 

\textbf{Main Results}:
The results for multivariate TS forecasting are outlined in Table \ref{tb2}, with the optimal results highlighted in {\bf bold} and the second-best results emphasized with \underline{underlined}. 
It can be found that, despite not utilizing MSE and MAE, DistPred achieves state-of-the-art performance across all datasets and prediction length configurations.
iTransformer and PatchTST stand out as the latest models acknowledged for their exceptional average performance.
Compared with them, the proposed DistPred demonstrates an average performance increase of {\bf 3.5\%} and {\bf 16.5\%}, respectively, achieving a substantial performance improvement.
We provide metrics, e.g., CRPS, QICE, PICP, for comparison by the future research community.
Note that for long time series, computing these metrics on the entire testset can be very time-consuming and may lead to out-of-memory issues. Therefore, we propose calculating these metrics for each batch and then averaging the results.

\subsection{Comparison with Conformal Prediction}

Since conformal prediction offers an alternative approach for distribution prediction, we compare DistPred with conformal prediction methods on the time series forecasting tasks, including CQR \citep{romano2019conformalized}, EnbPI \citep{xu2021conformal}, SPCI \citep{xu2023sequential}. 
Since PICP is the primary metric used by these methods, we adhere to their convention in this work.
As shown in Fig. \ref{fig_vs_cp}, DistPred achieves better performance than conformal prediction methods on all datasets and prediction lengths, with an average PICP improvement of 4.5\%. This further underscores the high competitiveness of DistPred.

\begin{table}
  \centering
  \caption{Stability experiments of loss function (Eq. \ref{eq_crps_effective}).}
  \resizebox{0.8\textwidth}{!}
  {
  \begin{tabular}{cccccc}
      \toprule
                 & 1\%       & 25\%       & 50\%          & 75\%          & 100\%      \\ \hline
      Train Loss & 0.21$\pm$0.04 & 0.174$\pm$0.03 & 0.165$\pm$0.01    & 0.162$\pm$0.01    & 0.16$\pm$0.007 \\ \hline
      Valid Loss & 0.17$\pm$0.05 & 0.155$\pm$0.04 &0.151$\pm$0.02     & 0.15$\pm$0.02     & 0.15$\pm$0.01    \\
      \bottomrule
  \end{tabular}
  }
  \label{tb_stable}
\end{table}

\begin{table}
  \centering
  \caption{KL-divergence, skewness, and kurtosis of the predictive response variable to the standard Gaussian on the ETTm2 datasets under input-96-predict-96 with  $K=100$.}
  \resizebox{0.7\textwidth}{!}
  {
  \begin{tabular}{cccccccc}
  \toprule
  Variates                        & 1     & 2    & 3     & 4    & 5     & 6     & 7     \\ \hline
  KL-Div                          & 0.11  & 0.1  & inf   & 0.13 & 0.03  & inf   & 0.05  \\
  Skewness                        & -0.42 & 0.41 & -0.47 & 0.16 & -0.03 & -0.26 & -0.72 \\
  Kurtosis                        & 0.77  & 0.09 & 1.93  & 0.31 & 0.85  & 1.95  & 1.77   \\
  \bottomrule
  \end{tabular}
  }
  \label{tb_dist}
\end{table}

\subsection{Stability of The Loss Function}

We included detailed discussions and experiments on the convergence and stability of the proposed method. As shown in Table \ref{tb_stable}, the experiments show that the training and valid loss converge smoothly across multiple runs, with no evidence of instability.

\begin{figure}
  \centering
  \centerline{\includegraphics[width=\columnwidth]{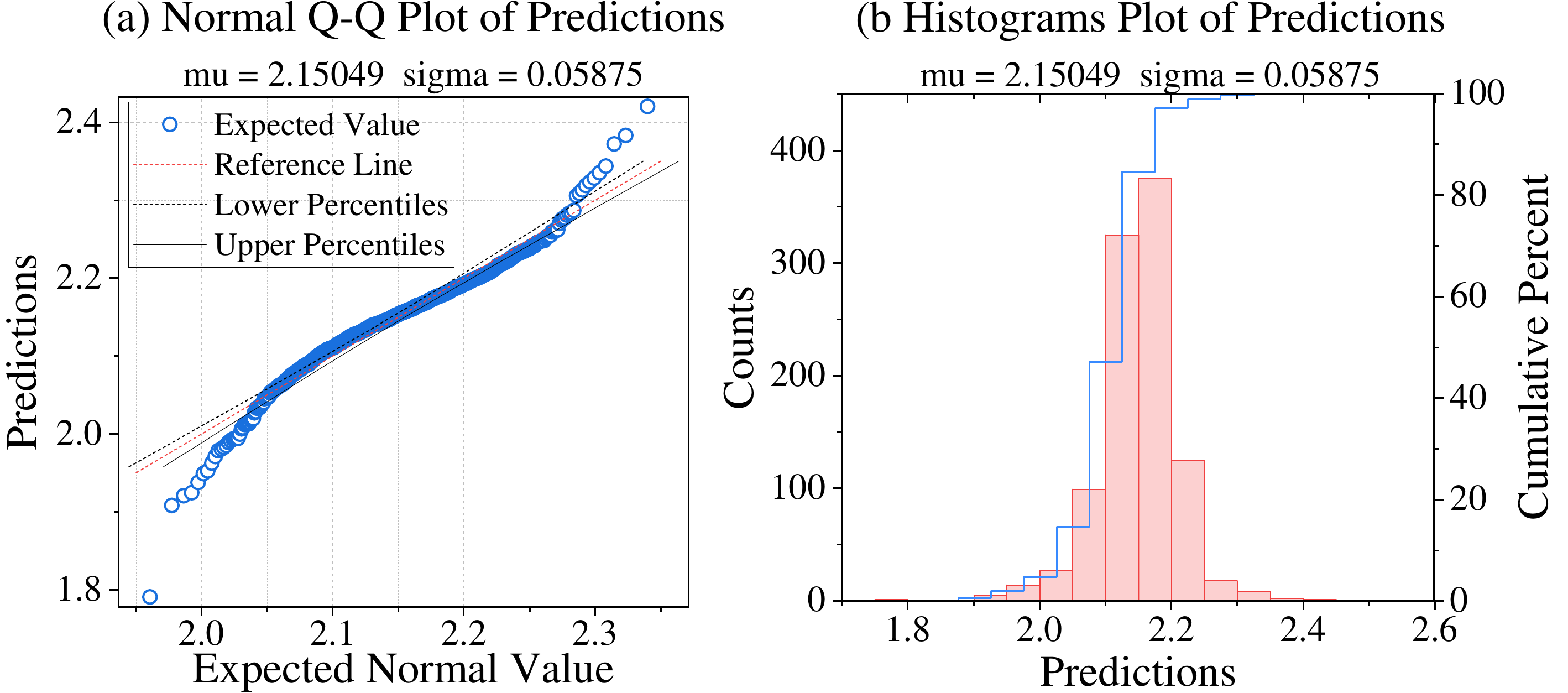}}
  \caption{QQ-plots and histograms of response variable.}
  \label{fig_qq_plot}
\end{figure}

\subsection{Visualization of The Predictive Distribution}

We conducted thorough statistical analyses, including QQ-plots, histograms, and metrics like KL-divergence, skewness, and kurtosis, to validate the empirical distribution of the predictive response variable.
KL-divergence, skewness, and kurtosis of the predictive response variable are shown in Table \ref{tb_dist}, and QQ-plots and histograms are shown in Fig. \ref{fig_qq_plot}.
These statistical analyses shows that the empirical distribution exhibits notable characteristics, such as skewness and heavy tails, etc., which are typical of many real-world datasets.

\begin{figure*}
  \centering
  \begin{minipage}{\textwidth}
  \centering
  \begin{subfigure}{0.49\textwidth}
    \centering
    \includegraphics[width=\textwidth]{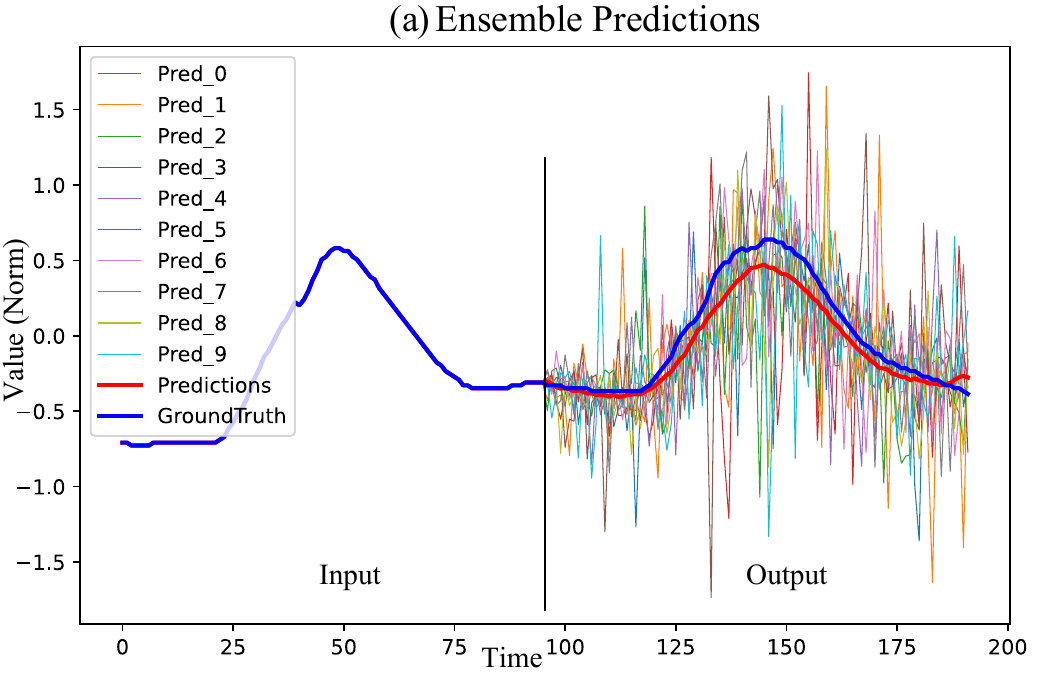}
  \end{subfigure}
  \begin{subfigure}{0.49\textwidth}
    \includegraphics[width=\textwidth]{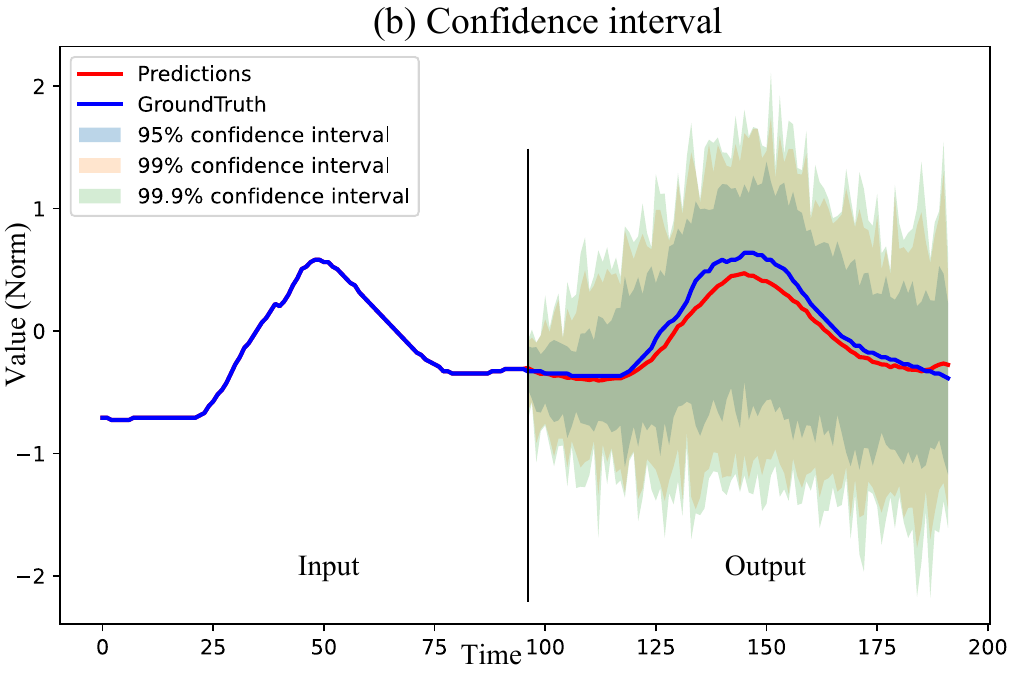}
  \end{subfigure}
  \caption{Visualization of the prediction results and the confidence intervals with setting input-96-predict-96 on the Ettm2 dataset. (a) The left subplot shows the ensemble of predictions with $K=100$. By utilizing subgraph (a), we can directly obtain the confidence intervals for subgraph (b), e.g, confidence intervals at 99\%, 99.5\% and 99.9\% levels. 
  }
  \label{fig_r6_vis}
\end{minipage}
\end{figure*}

Furthermore, the predictive distribution of DistPred is visualized in Fig. \ref{fig_r6_vis}. It can be observed that DistPred provides an ensemble of predictions (only the top 10 are presented in the left subplot).
Given all predictive ensemble values, the model can estimate the distribution of the response variable. Consequently, we can calculate confidence intervals at different levels, as shown in the right subplot of Fig. \ref{fig_r6_vis}.

\section{Related Work}
\label{app_relatedwork}

In supervised learning contexts, the endeavor to characterize the conditional distribution $p(y | x)$ beyond merely the conditional mean $\mathop{\mathbb{E}}[y | x]$ via deep neural networks has been a focal point of existing research efforts. These endeavors primarily concentrate on quantifying uncertainty, with several approaches having been proposed.

In regression and forecasting tasks, predicting the underlying distribution of the response variable is pivotal. Traditional approaches \citep{bishop1994mixture, greene2003econometric, salinas2020deepar, NEURIPS2020_NTK} often rely on the assumption that the response variable conforms to a prior distribution, with the goal of estimating specific statistics derived from this distribution. Commonly, these methods transform the challenge of distribution prediction and uncertainty quantification into the prediction of statistical parameters such as the mean and variance, under the presumption of a known continuous distribution. For example, MDNs \citep{bishop1994mixture} impose a predefined distribution—typically Gaussian—weighted by certain parameters to approximate the final distribution. Similarly, heteroscedastic regression \citep{greene2003econometric} predicts uncertainty by modeling the variability of residuals as a function of independent variables. DeepAR \citep{salinas2020deepar}, another notable approach, assumes a Gaussian distribution for the response variable, thereby leveraging the GaussianNLLLoss  \citep{Xie1994} to directly optimize its mean and variance. While these methods offer computational efficiency by simplifying predictions to statistical parameters, their reliance on strong distributional assumptions often limits their ability to capture the true underlying distribution, potentially leading to suboptimal performance.

Conformal prediction presents an alternative framework for distribution prediction, diverging from traditional parametric approaches. In their study, the authors in \citep{vovk2017nonparametric, vovk2018cross} introduced a random prediction system and proposed a nonparametric prediction method grounded in conformal assumptions. By integrating conformal prediction with quantile regression in \citep{romano2019conformalized, xu2021conformal, xu2023sequential}, they developed a method for constructing prediction intervals for the response variable.
However, the practical application of conformal prediction is not without limitations. Its effectiveness is often constrained by the assumption of exchangeability of residuals, which may not hold in all contexts, particularly in the presence of temporal dependencies. This limitation can lead to less reliable prediction intervals when applied to non-independent and identically distributed (non-i.i.d.) data, thereby challenging its robustness in real-world scenarios where data often exhibit complex dependencies.

Uncertainty quantification is a method that indirectly reflects the potential distribution of the response variable.
BNNs represent one such approach, aiming to capture such uncertainty by positing distributions over network parameters, thereby encapsulating the model's plausibility given the available data \citep{wierstra2015weight, hernandez2015probabilistic, gal2016dropout, vdropout, newelbo}. Another avenue is represented by \citet{alex2017uncertainty}, which not only addresses uncertainties in model parameters but also incorporates an additive noise term into the outputs to encompass uncertainties.
In parallel, ensemble-based methodologies \citep{lakshminarayanan2017simple, deepensemblesrecent} have emerged to address predictive uncertainty. These methods involve amalgamating multiple neural networks with stochastic outputs.
Furthermore, the neural processes' family \citep{np, cnp, anp, convcnp} has introduced a suite of models tailored to capturing predictive uncertainty in a manner that extends beyond the distribution of available data, particularly tailored for few-shot learning.

The aforementioned models have predominantly operated under the assumption of a parametric form in $p(y|x)$, typically adopting a Gaussian distribution or a mixture of Gaussians. They optimize network parameters by minimizing the negative log-likelihood of a Gaussian objective function.
In contrast, deep generative models are renowned for their capacity to model implicit distributions without relying on parametric distributional assumptions. However, only a sparse number of works have ventured into leveraging this capability to address regression tasks.
GAN-based models, as introduced by \citet{jointmatching} and \citet{wganjointmatching}, have emerged as one such endeavor, focusing on conditional density estimation and predictive uncertainty quantification. Additionally, \citet{han2022card} have proposed a diffusion-based model tailored for conditional density estimation.
Nevertheless, it is imperative to note that these models entail protracted training processes and computationally demanding inference procedures.

\section{Conclusion}

In this paper, we propose a novel method named DistPred, which is a distribution-free probabilistic inference approach for regression and forecasting tasks. We transform proper scoring rules that measure the discrepancy between the predicted distribution and the target distribution into a differentiable discrete form and use it as a loss function to train the model end-to-end. This allows the model to sample numerous samples in a single forward pass to estimate the potential distribution of the response variable. Experimental results demonstrate that DistPred outperforms existing methods, often by a considerable margin. We also extend time series forecasting from point estimation to distribution prediction and achieve state-of-the-art performance on multivariate and univariate time series forecasting tasks. In the future, we plan to extend DistPred to other tasks, such as classification and reinforcement learning.



\clearpage
\bibliographystyle{plainnat}
\bibliography{citations}


\clearpage
\appendix

\clearpage

\section{Dataset and Implementation}
\label{app_dataset}

\subsection{Commonly Used TS Datasets}

The information of the experiment datasets used in this paper are summarized as follows: (1) Electricity Transformer Temperature (ETT) dataset \cite{Zhou2021Informer}, which contains the data collected from two electricity transformers in two separated counties in China, including the load and the oil temperature recorded every 15 minutes (ETTm) or 1 hour (ETTh) between July 2016 and July 2018. (2) Electricity (ECL) dataset \footnote[1]{https://archive.ics.uci.edu/ml/datasets/ElectricityLoadDiagrams20112014} collects the hourly electricity consumption of 321 clients (each column) from 2012 to 2014. (3) Exchange \cite{lai2018modeling} records the current exchange of 8 different countries from 1990 to 2016. (4) Traffic dataset \footnote[2]{http://pems.dot.ca.gov} records the occupation rate of freeway system across State of California measured by 861 sensors. (5) Weather dataset \footnote[3]{https://www.bgc-jena.mpg.de/wetter} records every 10 minutes for 21 meteorological indicators in Germany throughout 2020. (6) Solar-Energy \cite{lai2018modeling} documents the solar power generation of 137 photovoltaic (PV) facilities in the year 2006, with data collected at 10-minute intervals. (7) The PEMS dataset \cite{liu2022scinet} comprises publicly available traffic network data from California, collected within 5-minute intervals and encompassing 358 attributes.
(8) Illness (ILI) dataset \footnote[4]{https://gis.cdc.gov/grasp/fluview/fluportaldashboard.html} describes the influenza-like illness patients in the United States between 2002 and 2021, which records the ratio of patients seen with illness and the total number of the patients. 
The detailed statistics information of the datasets is shown in Table \ref{tb1}, and the dataset information in terms of their size and number of features is summarized in Table \ref{tb_reg_uci_dim}.

\subsection{Implementation Details}
\label{app_implementation}

The model undergoes training utilizing the ADAM optimizer \cite{kingma2014adam} and minimizing the Mean Squared Error (MSE) loss function.
The training process is halted prematurely, typically within 10 epochs.
The DistPred architecture solely comprises the embedding layer and backbone architecture, devoid of any additional introduced hyperparameters.
During model validation, two evaluation metrics are employed: CRPS, QICE, PICP, MSE and MAE.
Given the potential competitive relationship between the two indicators, MSE and MAE, we use the average of the two ($\frac{MSE+MAE}{2}$) to evaluate the overall performance of the model.

\begin{table}
  \centering
  \caption{Details of the seven TS datasets. }
  \label{tb1}
  \resizebox{0.5\textwidth}{!}
  { \tiny
    \begin{tabular}{cccc}
      \toprule
       Dataset    & length  & features & frequency \\
      \midrule
      ETTh1       & 17,420  & 7       & 1h \\
      ETTh2       & 17,420  & 7       & 1h \\
      ETTm1       & 69,680  & 7       & 15m \\
      ETTm2       & 69,680  & 7       & 15m \\
      Electricity & 26,304  & 321     & 1h  \\
      Exchange    & 7,588   & 8       & 1d  \\
      Traffic     & 17,544  & 862     & 1h  \\
      Weather     & 52,696  & 21      & 10m \\
      Solar       & 52,560  & 137     & 10m  \\
      PEMS        & 26,208  & 358     & 5m  \\
      Illness     & 966     & 7       & 7d  \\
      \bottomrule
    \end{tabular}
  }
\end{table}

\begin{table*}
\caption{\label{tb_reg_uci_dim}Dataset size ($N$ observations, $P$ features) of UCI regression tasks.}
\begin{center}
\resizebox{\textwidth}{!}{
\begin{tabular}{@{}c|cccccccccc@{}}
\toprule
Dataset  & Boston      & Concrete     & Energy    & Kin8nm       & Naval          & Power        & Protein       & Wine          & Yacht     & Year            \\ \midrule
$(N, P)$ & $(506, 13)$ & $(1030,8)$ & $(768,8)$ & $(8192,8)$ & $(11,934,16)$ & $(9568,4)$ & $(45,730,9)$ & $(1599,11)$ & $(308,6)$ & $(515,345,90)$     \\ \bottomrule
\end{tabular}}
\end{center}
\end{table*}

\section{Comparisons of Model Efficiency in Time Series Forecasting}

\begin{figure}
  \centering
  \centerline{\includegraphics[width=0.8\textwidth]{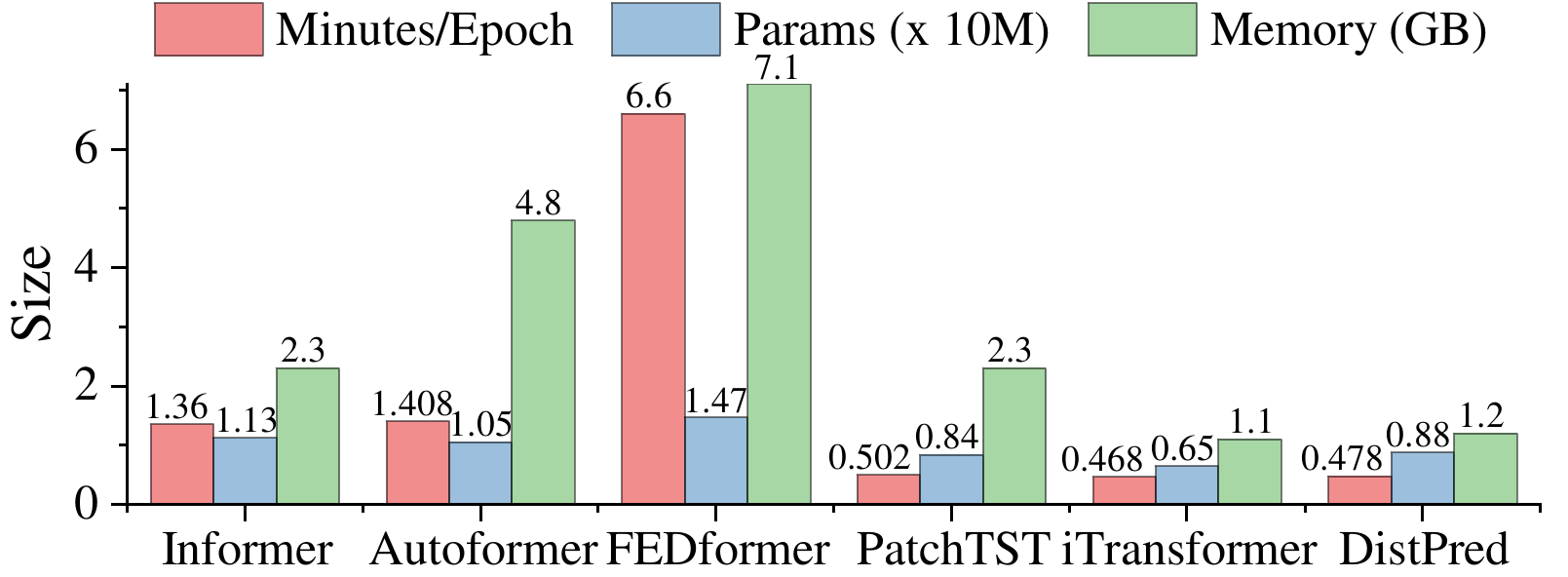}}
  \caption{QQ-plots and histograms of response variable.}
  \label{fig_ts_runtime}
\end{figure}

We evaluate the inference time and size of DistPred (mean only) in comparison to other Transformer-based models, as shown in Fig. \ref{fig_ts_runtime}.
It can be observed that DistPred's runtime, model size, and memory usage do not significantly increase compared to other models.

\section{Distribution Free v.s. Distribution Related}

If we assume that the response variable follows a continuous distribution, as done in \citep{salinas2020deepar} where $\hat{y}$ is assumed to be followed Gaussian distribution, we can provide an analytical formula for the Gaussian likelihood. Specifically, if $\hat{y} \sim \mathcal{N}(\mu, \sigma)$, then we can parametrize the Gaussian likelihood using its mean and standard deviation,
\begin{equation}
  \mathcal{L}_G(\hat{y}|\mu, \sigma) = (2\pi\sigma^2)^{-\frac{1}{2}}exp(-(\hat{y}-\mu)^2/(2\sigma^2) ) . \label{eq_gaussian}
\end{equation} 

Then, we can train DistPred utilizing Eq. \ref{eq_gaussian} as the loss function. As shown in Table \ref{tb_dp_gauss}, DistPred exclusively provides the mean and variance of the response variable.

\begin{table*}
  \centering
  \caption{Comparison of DistPred trained with Gaussia distribution.}
  \resizebox{0.8\textwidth}{!}
  {
  \begin{tabular}{c|ccc|ccc|ccc|ccc}
    \toprule
             & \multicolumn{3}{c|}{Ettm2} & \multicolumn{3}{c|}{Traffic} & \multicolumn{3}{c|}{Exchange} & \multicolumn{3}{c}{Weather} \\
             \midrule
             & CRPS     & MSE   & MAE    & CRPS     & MSE     & MAE    & CRPS     & MSE     & MAE     & CRPS     & MSE     & MAE    \\
             \midrule
  DistPred   & 0.189    & 0.12  & 0.256  & 0.161    & 0.137   & 0.216  & 0.343    & 0.426   & 0.456   & 0.023    & 0.002   & 0.028  \\
  DistPred-G & 4391.80  & 7.61  & 1.73   & 2329.05  & 276.51  & 6.35   & 177.92   & 16.52   & 1.72    & 229.58   & 2.27    & 0.17  \\
  \bottomrule
  \end{tabular}
  }
  \label{tb_dp_gauss}
\end{table*}

\section{DistPred for Time Series with Missing Values}
\label{app_ts_miss}

The proposed DistPred is equipped to handle time series with missing values scenarios, as both imputation and forecasting are done in a similar manner. Our experiments have demonstrated that DistPred remains SOTA even when a substantial portion of the data is missing. Specifically, in cases where 80\% of the PhysioNet2012 data points were absent, the model still maintained competitive performance, as shown in Table \ref{tb_ts_miss}.

\begin{table*}
  \centering
  \caption{Comparison of DistPred with other SOTA models when time series with missing values.}
  \resizebox{\textwidth}{!}
  {
    \Large
    \begin{tabular}{cccccccccc}
    \toprule
    Model & M-RNN       & GP-VAE      & BRITS       & USGAN       & CSDI        & TimesNet    & Transformer & SAITS       & DistPred    \\ \midrule
    MSE   & 0.864$\pm$0.002 & 0.433$\pm$0.011 & 0.325$\pm$0.002 & 0.306$\pm$0.001 & 0.260$\pm$0.05  & 0.272$\pm$0.006 & 0.225$\pm$0.002 & 0.218$\pm$0.002 & \bf 0.204$\pm$0.002 \\
    MAE   & 0.674$\pm$0.001 & 0.4$\pm$0.007   & 0.246$\pm$0.001 & 0.25$\pm$0.001  & 0.211$\pm$0.003 & 0.266$\pm$0.007 & 0.209$\pm$0.002 & 0.202$\pm$0.002 & \bf 0.196$\pm$0.002 \\ 
    \bottomrule
    \end{tabular}
}
\label{tb_ts_miss}
\end{table*}

\section{Univariate Time Series Forecasting}
\label{app_univariate}

The full results for univariate TS forecasting are presented in Table \ref{tb4}. 
As other models, e.g., iTransformer \cite{liu2023itransformer} and PatchTST \cite{nie2022time}do not offer performance information for all prediction lengths, we compare our method with those that provide comprehensive performance analysis, including  FEDformer \cite{zhou2022fedformer}, Autoformer \cite{wu2021autoformer}, Informer \cite{Zhou2021Informer}, LogTrans \cite{li2019enhancing} and Reformer \cite{Kitaev2020Reformer}.
This reaffirms the effectiveness of DistPred.

\begin{table*}
  \centering
  \caption{Univariate time series forecasting results on benchmark datasets.}
  \label{tb4}
  \resizebox{\textwidth}{!}
  {
  
  \begin{tabular}{cc|ccccc| cc| cc| cc| cc| cc}
    \toprule
    \multicolumn{2}{c}{Model}          & \multicolumn{5}{c}{DistPred-96}  & \multicolumn{2}{c}{FEDformer-96} & \multicolumn{2}{c}{Autoformer-96} & \multicolumn{2}{c}{Informer-96} & \multicolumn{2}{c}{LogTrans-96} & \multicolumn{2}{c}{Reformer-96} \\
    \toprule
                              & Length & CRPS & QICE              & PICP             & MSE               & MAE               & MSE             & MAE            & MSE             & MAE             & MSE            & MAE            & MSE            & MAE            & MSE            & MAE            \\ \toprule
    
    \multirow{5}{*}{\rotatebox{90}{ETTh1}}       
      & 96  & {0.134}   & {6.341}           & { 96.936}             &{ \bf0.057}             & {\bf0.182}             & 0.079           & 0.215          &\underline{0.071}           &\underline {0.206}           & 0.193          & 0.377          & 0.283          & 0.468          & 0.532          & 0.569          \\
      & 192 & {0.152}  & {3.766}           & { 97.479}             & { \bf0.073}             & {\bf0.207}            & \underline{0.104}           &\underline {0.245}          & 0.114           & 0.262           & 0.217          & 0.395          & 0.234          & 0.409          & 0.568          & 0.575          \\
      & 336 &{0.162}   & {3.762}           &{ 98.071}             &{ \bf0.080}             & {\bf0.221}             & 0.119           & 0.270           & \underline{0.107}           &\underline {0.258}           & 0.202          & 0.381          & 0.386          & 0.546          & 0.635          & 0.589          \\
      & 720 &{0.164}   & {4.176}           & { 88.437}             & { \bf0.082}             &{\bf0.226}             & 0.142           & 0.299          & \underline{0.126}          &\underline {0.283}           & 0.183          & 0.355          & 0.475          & 0.628          & 0.762          & 0.666          \\ \cline{2-17}
      & Avg &{0.153}  & {4.505}           & { 95.235}             & {\bf0.073}             &{\bf0.209}             & 0.111           & 0.257          & \underline{0.105}           &\underline {0.252}           & 0.199          & 0.377          & 0.345          & 0.513          & 0.624          & 0.600          \\ \toprule
    
    \multirow{5}{*}{\rotatebox{90}{ETTh2}}       
      & 96  &{0.203} & 7.511           & {81.780}             &\underline {0.129}             & \underline { 0.276}             & { \bf0.128}           & {\bf0.271}          & 0.153           & 0.306           & 0.213          & 0.373          & 0.217          & 0.379          & 1.411          & 0.838          \\ 
      & 192 &{ 0.240}  & { 4.164}           & { 80.855}             & {\bf0.176}             & {\bf0.327}              &\underline{ 0.185}           &\underline {0.33}           & 0.204           & 0.351           & 0.227          & 0.387          & 0.281          & 0.429          & 5.658          & 1.671          \\
      & 336 &{0.281}  & {5.482}           & { 87.817}             & \underline{ 0.234}             &{\bf0.348}             & {\bf0.231}           &\underline {0.378}          & 0.246           & 0.389           & 0.242          & 0.401          & 0.293          & 0.437          & 4.777          & 1.582          \\
      & 720 &{0.274} & {5.820}           &  89.181             &  \underline{0.219}             &{\bf0.379}              & 0.278           & 0.42           & 0.268           & 0.409           & 0.291          & 0.439          & {\bf0.218}          & \underline { 0.387}          & 2.042          & 1.039          \\ \cline{2-17}
      & Avg &{0.250} & {5.744}           &{ 84.908}             & { \bf0.190}             &{\bf0.341}           &\underline{0.206}         &\underline {0.350}          & 0.218           & 0.364           & 0.243          & 0.400          & 0.252          & 0.408          & 3.472          & 1.283          \\ \toprule
    
    \multirow{5}{*}{\rotatebox{90}{ETTm1}}       
      & 96  &{0.095} & {6.492}           &{ 88.992}             & { \bf0.029}             & {\bf0.126}             &\underline{0.033}           & \underline {0.140}           & 0.056           & 0.183           & 0.109          & 0.277          & 0.049          & 0.171          & 0.296          & 0.355          \\ 
      & 192 &{0.117}  & {4.496}           & {92.619}             & { \bf0.044}             &  {\bf0.158}            &\underline{0.058}           & \underline {0.186}          & 0.081           & 0.216           & 0.151          & 0.310           & 0.157          & 0.317          & 0.429          & 0.474          \\
      & 336 &{0.137}  & {5.488}           & { 92.264}              & {\bf0.058}             & {\bf0.186}             & 0.084           &  0.231          &\underline{0.076}           &\underline {0.218}           & 0.427          & 0.591          & 0.289          & 0.459          & 0.585          & 0.583          \\
      & 720 &{0.163}  & {3.541}           & { 95.713}             & { \bf0.080}             & {\bf0.218}             & \underline{0.102}           & \underline {0.250}           & 0.110            & 0.267           & 0.438          & 0.586          & 0.430           & 0.579          & 0.782          & 0.73           \\ \cline{2-17}
      & Avg &{0.128} & {5.004}           & { 92.397}             & { \bf0.053}             & {\bf0.172}             & \underline{0.069}          & \underline {0.202}          & 0.081           & 0.221           & 0.281          & 0.441          & 0.231          & 0.382          & 0.523          & 0.536          \\ \toprule
    
    \multirow{5}{*}{\rotatebox{90}{ETTm2}}       
      & 96  &{0.133} & {9.595}          & {69.279}             & \underline{0.064}             & {\bf0.180}             & {\bf0.063}           & \underline {0.189}          & 0.065           & \underline {0.189}           & 0.08           & 0.217          & 0.075          & 0.208          & 0.077          & 0.214          \\ 
      & 192 & {0.173} & {7.392}          &{75.604}             & {\bf0.099}             & {\bf0.233}             & \underline{0.110}           & \underline {0.252}          & 0.118           & 0.256           & 0.112          & 0.259          & 0.129          & 0.275          & 0.138          & 0.290           \\
      & 336 &{0.203} & {8.833}          & {82.711}             & {\bf0.133}             & {\bf0.277}            & \underline{0.147}           & \underline {0.301}          & 0.154           & 0.305           & 0.166          & 0.314          & 0.154          & 0.302          & 0.160          & 0.313          \\
      & 720 &{0.248}  & 5.720            & {90.470}             & {0.185}             & \underline {0.333}             & 0.219           & 0.368          & 0.182           & 0.335           & 0.228      & 0.380           & {\bf0.160}          & {\bf0.322}         & \underline{0.168}          & 0.334          \\ \cline{2-17}
      & Avg &{0.189} &{7.885}           & {79.516}             & {\bf0.120}             & {\bf0.256}             & 0.135           & 0.278          & \underline{0.130}           & \underline {0.271}           & 0.147          & 0.293          & \underline{0.130}          & 0.277          & 0.136          & 0.288          \\ \toprule
    
    \multirow{5}{*}{\rotatebox{90}{Traffic}}     
      & 96  & {0.155}            & 4.245           & 25.012             & {\bf0.132}             & {\bf0.209}             & \underline{0.170}           & \underline{0.263}          & 0.246           & 0.346           & 0.257          & 0.353          & 0.226          & 0.317          & 0.313          & 0.383          \\ 
      & 192 & {0.158}            & 3.596         & 25.458             & {\bf0.136}             & {\bf0.213}             & \underline{0.173}           & \underline{0.265}          & 0.266           & 0.37            & 0.299          & 0.376          & 0.314          & 0.408          & 0.386          & 0.453          \\
      & 336 & {0.171}             & 3.458           & 25.146             & {\bf0.134}             & {\bf0.213}             & \underline{0.178}           & \underline{0.266}          & 0.263           & 0.371           & 0.312          & 0.387          & 0.387          & 0.453          & 0.423          & 0.468          \\
      & 720 & {0.135}            & 3.434           & 24.926             & {\bf0.146}            & {\bf0.228}             & \underline{0.187}           & \underline{0.286}          & 0.269           & 0.372           & 0.366          & 0.436          & 0.437          & 0.491          & 0.378          & 0.433          \\ \cline{2-17}
      & Avg & {0.161}            & 3.683           & 25.136             & {\bf0.137}            &{\bf0.216}             & \underline{0.177}           & \underline{0.220}          & 0.261           & 0.365           & 0.309          & 0.388          & 0.341          & 0.417          & 0.375          & 0.434          \\ \toprule
    
    \multirow{5}{*}{\rotatebox{90}{Electricity}} 
      & 96  & 0.266            & 5.619           & 51.949             & {\bf0.257}             & {\bf0.366}             & 0.262           & 0.378          & 0.341           & 0.438           & \underline{0.258}          &\underline{0.367}          & 0.288          & 0.393          & 0.275          & 0.379          \\
      & 192 & {0.276}            & 5.018           & 65.785             & {\bf0.284}            & {\bf0.376}             & 0.316           & 0.410          & 0.345           & 0.428           & \underline{0.285}          & \underline{0.388}          & 0.432          & 0.483          & 0.304          & 0.402          \\
      & 336 & {0.317}           & 4.947           & 62.019             & 0.400               & 0.439              & \underline{0.361}           & \underline{0.445}          & 0.406           & 0.470          & {\bf0.336}          & {\bf0.423}          & 0.430          & 0.483          & 0.37           & 0.448          \\
      & 720 & {0.335}            & 5.191           & 67.705             & {\bf0.409}             & {\bf0.460}            & \underline{0.448}           & \underline{0.501}          & 0.565           & 0.581           & 0.607          & 0.599          & 0.491          & 0.531          & 0.46           & 0.511          \\ \cline{2-17}
      & Avg & {0.299}            & 5.191           & 61.865             & {\bf0.337}             & {\bf0.410}             & \underline{0.347}           & \underline{0.434}          & 0.414           & 0.479           & 0.372          & 0.444          & 0.410          & 0.473          & 0.352          & 0.435          \\ \toprule
    
    \multirow{5}{*}{\rotatebox{90}{Weather}}     
      & 96  & 0.019           & 10.732          & 61.824            & {\bf0.0012}              & {\bf0.024}              & \underline{0.0035}          & 0.046          & 0.0110          & 0.081           & 0.004          & \underline{0.044}          & 0.0046         & 0.052          & 0.012          & 0.087          \\
      & 192 & 0.022           & 10.463          & 55.581            & {\bf0.0013}            & {\bf0.027}            & 0.0054          & 0.059          & 0.0075          & 0.067           & \underline{0.002}          & \underline{0.040}          & 0.006          & 0.060          & 0.0098         & 0.044          \\
      & 336 & 0.023           & 10.271          & 59.075            & {\bf0.0021}            & {\bf0.024}            & 0.008           & 0.072          & 0.0063          & 0.062           & \underline{0.004}          & \underline{0.049}          & 0.006          & 0.054          & 0.013          & 0.100          \\
      & 720 & 0.027            & 10.515         & 47.492             & {\bf0.0023}            & {\bf0.033}            & 0.015           & 0.091          & 0.0085          & 0.070           & \underline{0.003}          & \underline{0.042}          & 0.007          & 0.059          & 0.011          & 0.083          \\ \cline{2-17}
      & Avg & 0.023           & 10.495          & 55.993            & {\bf0.0023}            & {\bf0.028}            & 0.008           & 0.067          & 0.0083          & 0.0700          & \underline{0.0033}         & \underline{0.0438}         & 0.0059         & 0.0563         & 0.0115         & 0.0785         \\ \toprule
    
    \multirow{5}{*}{\rotatebox{90}{Exchange}}    
      & 96  & 0.182            & 6.99           & 94.678             & {\bf0.110}             & {\bf0.241}             & \underline{0.131}           & \underline{0.284}          & 0.241           & 0.387           & 1.327          & 0.944          & 0.237          & 0.377          & 0.298          & 0.444          \\
      & 192 & {0.263}              & 10.391           & 82.484             & {\bf0.204}             &  {\bf0.338}             &  \underline{0.277}           & 0.420          & 0.300           & \underline{0.369}           & 1.258          & 0.924          & 0.738          & 0.619          & 0.777          & 0.719          \\
      & 336 & {0.376}              & 10.778           & 87.861             & {\bf0.401}             & {\bf0.482}             & \underline{0.426}           & \underline{0.511}          & 0.509           & 0.524           & 2.179          & 1.296          & 2.018          & 1.0700         & 1.833          & 1.128          \\
      & 720 & {0.552}             & 9.526           & 81.032             & {\bf0.991}             & {\bf0.763}              & \underline{1.162}           & \underline{0.832}          & 1.260           & 0.867           & 1.28           & 0.953          & 2.405          & 1.175          & 1.203          & 0.956          \\ \cline{2-17}
      & Avg & {0.343}            & 9.421           & 86.514             & {\bf0.426}             & {\bf0.456}             & \underline{0.499}           & \underline{0.512}          & 0.578           & 0.537           & 1.511          & 1.029          & 1.350          & 0.810          & 1.028          & 0.812          \\ \toprule
    
      \multicolumn{2}{c}{ $1^{\text{st}}$ Count} 
      &- & -         &      -     & {\bf 34}                     & {\bf37}     &  \underline{3}                  & 1               & 0             & 0               & 1               & 1               & 2             & 1             & 0               & 0             \\
    \bottomrule
    \end{tabular}
}
\end{table*}

\end{document}